\documentclass[sigconf]{acmart}
\AtBeginDocument{%
  }

\copyrightyear{2025}
\acmYear{2025}
\setcopyright{acmlicensed}
\acmConference[CIKM '25] {Proceedings of the 34th ACM International Conference on Information and Knowledge Management}{ November 10--14, 2025}{Seoul, Republic of Korea.}
\acmBooktitle{Proceedings of the 34th ACM International Conference on Information and Knowledge Management (CIKM '25), November 10--14, 2025, Seoul, Republic of Korea}
\acmISBN{979-8-4007-2040-6/2025/11}
\acmDOI{10.1145/3746252.3761191}

\usepackage{subcaption}
\usepackage{multirow}
\usepackage[ruled,vlined]{algorithm2e}

\begin{document}

\title{Revisiting Long-Tailed Learning: Insights from an Architectural Perspective}


\author{Yuhan Pan}
\affiliation{%
  \institution{University of Science and Technology of China}
  \city{Hefei}
  \country{China}
}
\email{yuhanpan@mail.ustc.edu.cn}

\author{Yanan Sun}
\affiliation{%
 \institution{Sichuan University}
 \city{Chengdu}
 \country{China}}
\email{ysun@scu.edu.cn}

\author{Wei Gong}
\authornote{Corresponding author.}
\affiliation{%
  \institution{University of Science and Technology of China}
  \city{Hefei}
  \country{China}
}
\email{weigong@ustc.edu.cn}

\renewcommand{\shortauthors}{Yuhan Pan, Yanan Sun, \& Wei Gong}

\begin{abstract}
Long-Tailed (LT) recognition has been widely studied to tackle the challenge of imbalanced data distributions in real-world applications. However, the design of neural architectures for LT settings has received limited attention, despite evidence showing that architecture choices can substantially affect performance. This paper aims to bridge the gap between LT challenges and neural network design by providing an in-depth analysis of how various architectures influence LT performance. Specifically, we systematically examine the effects of key network components on LT handling, such as topology, convolutions, and activation functions. Based on these observations, we propose two convolutional operations optimized for improved performance. Recognizing that operation interactions are also crucial to network effectiveness, we apply Neural Architecture Search (NAS) to facilitate efficient exploration. We propose LT-DARTS, a NAS method with a novel search space and search strategy specifically designed for LT data. Experimental results demonstrate that our approach consistently outperforms existing architectures across multiple LT datasets, achieving parameter-efficient, state-of-the-art results when integrated with current LT methods.
\end{abstract}

\ccsdesc[500]{Computing methodologies~Neural networks}
\keywords{Neural Architecture Search, Image Classification, Long-Tailed Learning}


\maketitle

\section{Introduction}
\label{sec:intro}

Although deep neural networks have been proven effective in solving many problems~\cite{10577272, 10388040, 9796782, 10138664}, their robustness under imbalanced data distributions still requires further investigation. In reality, data distributions in many practical scenarios are Long-Tailed (LT) \cite{Buda2017ASS, wang2018bravo, liu2023graph}, characterized by a few head classes with abundant samples and many tail classes with limited samples. It has been proved that conventional deep learning methods tend to exhibit sub-optimal performance in such a context. To address this challenge, the community has proposed numerous approaches including data augmentation~\cite{Li2021MetaSAugMS, Hong2022SAFASF, ji2024fedfixer, Ahn2023CUDACO}, class re-sampling strategies~\cite{Liu2021GistNetAG, Bai2023OnTE}, decoupling learning~\cite{Kang2019DecouplingRA, zhou2023ccsam} and so on. Nevertheless, most existing methods pay insufficient attention to neural architecture design, which is crucial for deep models to achieve optimal performance.

\begin{figure}[t]
    \centering
    \includegraphics[width=0.45\textwidth]{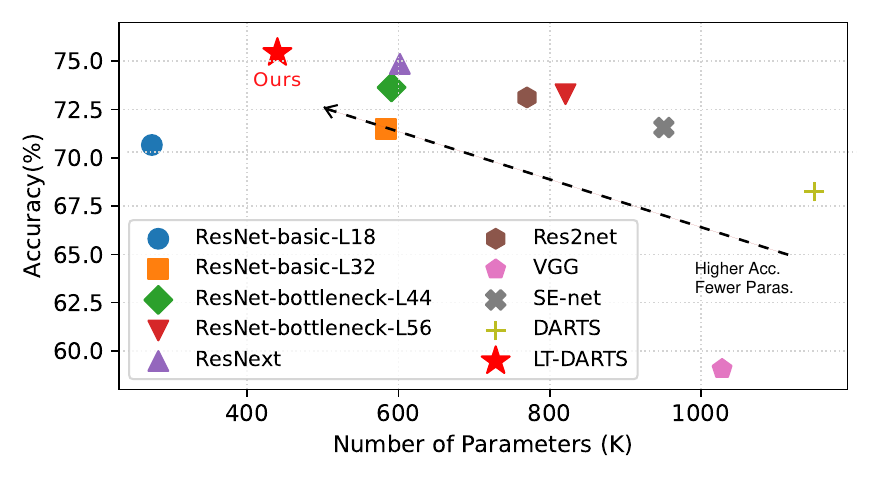}
    \caption{The performance of different architectures on the long-tailed CIFAR-10-LT dataset. The direction of black dashed lines indicates better architectures, as they achieve higher accuracy with fewer architectural parameters.}
    \label{fig:first}
\end{figure}

It is widely recognized that meticulous architectural design contributes to improved performance on balanced data distribution, and this consensus unsurprisingly extends to LT data distribution. As depicted in Fig.~\ref{fig:first}, we conduct extensive training on several architectures and report their parameter sizes and performance. It can be observed that different backbones lead to various performances and an optimal architecture may achieve superior performance even with fewer parameters, which validates the feasibility of approaching the LT problem from an architectural perspective.

\begin{figure}[tbp]
    \centering
    \begin{subfigure}[b]{0.23\textwidth}
        \centering
        \includegraphics[width=\textwidth]{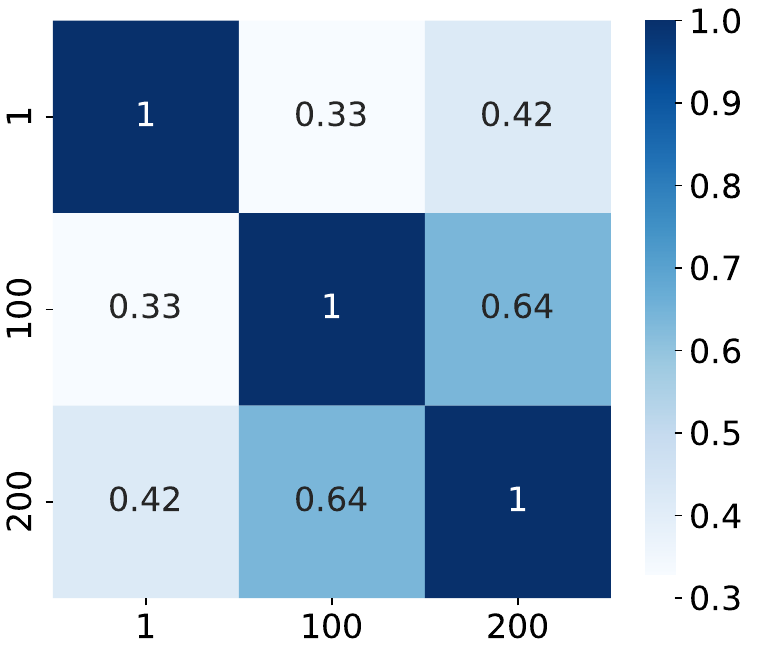}
        \caption{CIFAR-10-LT}
    \end{subfigure}
    \hfill
    \begin{subfigure}[b]{0.23\textwidth}
        \centering
        \includegraphics[width=\textwidth]{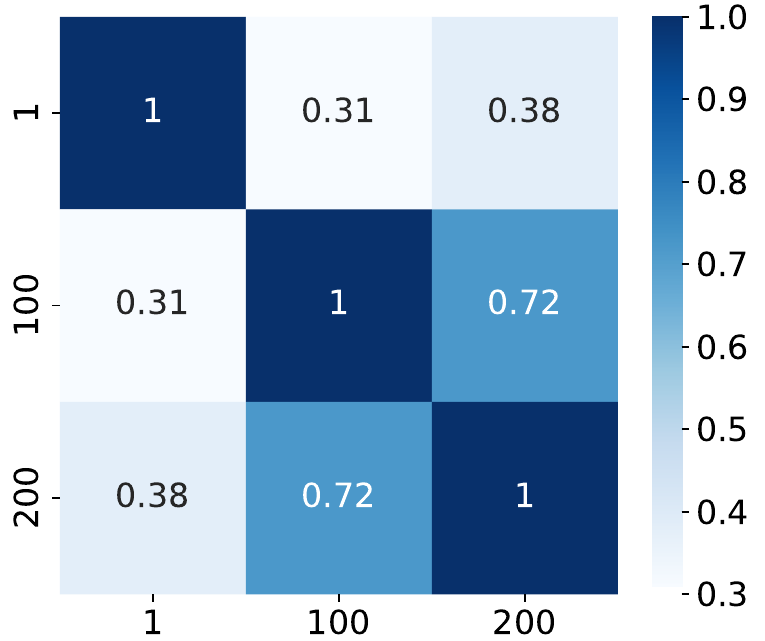}
        \caption{CIFAR-100-LT}
    \end{subfigure}
    \caption{Heatmap of Kendall's Tau correlation under different imbalance factors. A lower tau (lighter color) indicates weaker performance correlation.}
    \label{fig:rank}
\end{figure}

While neural architectures have been extensively evaluated on balanced datasets, it remains unclear \textit{whether these architectures generalize effectively to LT distributions}. To explore this, we conduct a preliminary study. We randomly sample 10 architectures from NAS-Bench-201~\cite{Dong2020NASBench201ET}, a benchmark comprising 15,625 candidate networks. Each architecture is evaluated under three class imbalance settings with imbalance factors $\rho \in \{1, 100, 200\}$, where $\rho = n_{max} / n_{min}$ denotes the ratio of the largest to smallest class sizes. For each setting, we record accuracy-based rankings of the architectures. We then compute Kendall’s Tau correlation between the rankings across different imbalance levels (Figure~\ref{fig:rank}). Results show weak correlations, particularly between balanced and highly imbalanced settings, indicating that architectures performing well under balanced conditions may not retain their performance in LT scenarios. These findings suggest that architecture performance is not invariant to class distribution, motivating the need to design and search for architectures explicitly suited to long-tailed data.

To this end, we first conduct a systematic analysis of key architectural components under LT data, including network topology, activation functions, convolutional operations, and normalization methods. Our study reveals several design patterns that consistently improve performance in LT settings. Building on these insights, we introduce two new convolutional operations, named LT-AggConv and LT-HierConv, tailored for LT scenarios. Since the arrangement and interaction of operations also play a critical role in model performance, we further propose LT-DARTS, a neural architecture search (NAS) algorithm with an LT-aware search space and an unbiased search strategy, enabling efficient discovery of effective architectures.
Extensive experiments show that our method mitigates LT issues at the architectural level. Moreover, it is orthogonal to existing approaches and can be easily combined with them to achieve state-of-the-art results. Our contributions can be concluded as the following:

\begin{itemize}
    \item We bridge the gap between neural architecture and LT. Through extensive experimentation, we identify the architectural components that perform better in LT scenarios and propose LT-AggConv and LT-HierConv, two convolutions specifically designed for LT.

    \item We propose LT-DARTS, including a search space better suited for LT and a de-biasing search strategy, to explore the use of NAS in developing an LT-friendly architecture.

    \item  Extensive experiments show that the architectures we discover outperform crafted designed architectures. And it can seamlessly integrate with existing solutions tailored to LT, further enhancing model performance. Our research provides a complementary perspective for the LT community.
\end{itemize}
\section{Related Work}
\subsection{Deep Long-tailed Learning}
Recently much effort has been devoted to deep long-tailed recognition. A branch of works~\cite{Cao2019LearningID, Zhou2019BBNBN, Hong2020DisentanglingLD, Zhang2021DistributionAA} tackle the LT problem from the perspective of class re-balance, proposing novel sampling strategies and loss functions to effectively re-weight imbalanced samples. Meanwhile, data augmentation is also introduced to mitigate this issue~\cite{Park2021TheMC, Ahn2023CUDACO}. They explore how to enhance the representation of minority classes through data processing.
Moreover, some studies~\cite{Kang2019DecouplingRA, Zhong2020ImprovingCF, Wu2021AdversarialRU} decouple the feature representation and classifier, improving each module independently, which effectively enhanced the performance of the model on LT data. However, existing methods have paid little attention to the influence of network architecture on the long-tailed problem, as most of them directly adopt popular backbone networks such as ResNet~\cite{He2015DeepRL} for feature extraction. This paper takes a different approach by focusing on the design of the network, providing a complementary perspective for LT.
\subsection{Differentiable Architecture Search}
As one of the representative methods of NAS, Differentiable Architecture Search (DARTS) has garnered significant attention in recent years because of its powerful performance and reduced computational requirements. Liu et al. pioneer the introduction of DARTS~\cite{Liu2018DARTSDA}, allowing an efficient search of the architecture using gradient descent. Subsequently, a series of variations have been proposed to optimize the vanilla version. PC-DARTS \cite{Xu2019PCDARTSPC} achieves higher efficiency by randomly sampling a subset of channels for operation search. R-DARTS \cite{Zela2019UnderstandingAR} proposes an early stopping criterion to make the searched model more robust. Fair-DARTS \cite{Chu2019FairDE} relaxes the exclusive competition between skip-connections and achieves higher performance. Although these works have improved DARTS to some extent, their settings are still based on uniform data distribution. In this paper, we introduce DARTS to LT and propose targeted improvements to address its limitations under LT, allowing it to search for architectures with enhanced performance.

\begin{figure*}[t]
    \centering
    \includegraphics[width=0.93\textwidth]{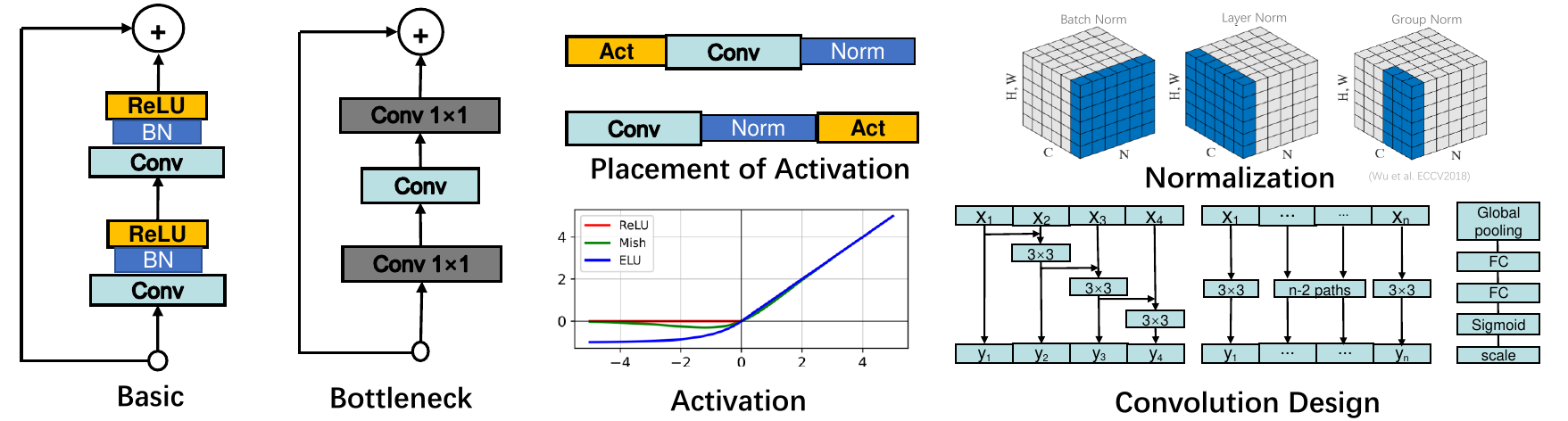}
    \caption{ An overview of the architectural properties we explore. ``Basic'' represents the foundational architecture. ``Bottleneck'' refers to an improved topology, with the activation functions and normalization methods for each layer omitted for simplicity. Additionally, we investigate the design of convolution, placement of activation, normalization, and activation.}
    \label{fig:explore_com}
\end{figure*}

\section{Investigation of Components for LT}
To design a LT-friendly architecture, we first conduct extensive experiments to explore the performance of architectural components under LT distributions. As shown in Fig.~\ref{fig:explore_com}, our explorations cover various aspects, including topology, the specific design of convolutions, the selection and placement of activation functions, and different normalization methods. Based on experimental observations, we distill the key components suited for LT data and propose two new convolutional architectures: LT-AggConv and LT-HierConv.

\subsection{Preliminaries} \label{pre}
We initially introduce the experimental setup for exploring LT-friendly networks and the evaluation metric.

\textbf{Basic architecture.} The current LT community universally uses ResNet series networks as the backbone for feature extraction~\cite{Hong2020DisentanglingLD, Ren2020BalancedMF, Wang2021RSGAS, Zhong2020ImprovingCF}. Without loss of generality, our exploration is also based on ResNet~\cite{He2015DeepRL}. The basic architecture consists of three network layers, each comprising several stacked blocks. The block design of the baseline follows the ResNet basic block structure, where each block is composed of two identical convolution-BN-ReLU modules and a shortcut connection. 

\textbf{Training.} We employ the standard cross-entropy loss and set the batch size to 128. The optimization uses the SGD optimizer with a learning rate of 0.1, a momentum of 0.9, and a weight decay 2e-4. These settings are chosen to ensure an appropriate balance between effective learning dynamics and regularization.

\textbf{Evaluation.} We evaluate the performance of individual components on the CIFAR-10-LT and CIFAR-100-LT datasets. These two datasets are variations of the vanilla CIFAR dataset, sampled with different quantities across classes to create LT datasets.

\begin{figure}[tbp]
    \centering
    \begin{subfigure}[b]{0.23\textwidth}
        \centering
        \includegraphics[width=\textwidth]{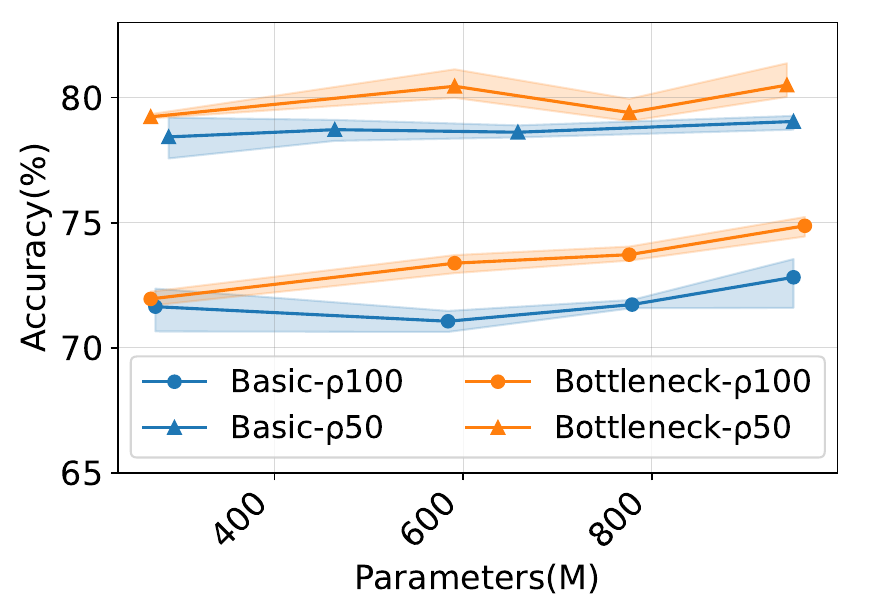}
        \caption{CIFAR-10-LT}
        \label{fig:basic_bottle_10}
    \end{subfigure}
    \hfill
    \begin{subfigure}[b]{0.23\textwidth}
        \centering
        \includegraphics[width=\textwidth]{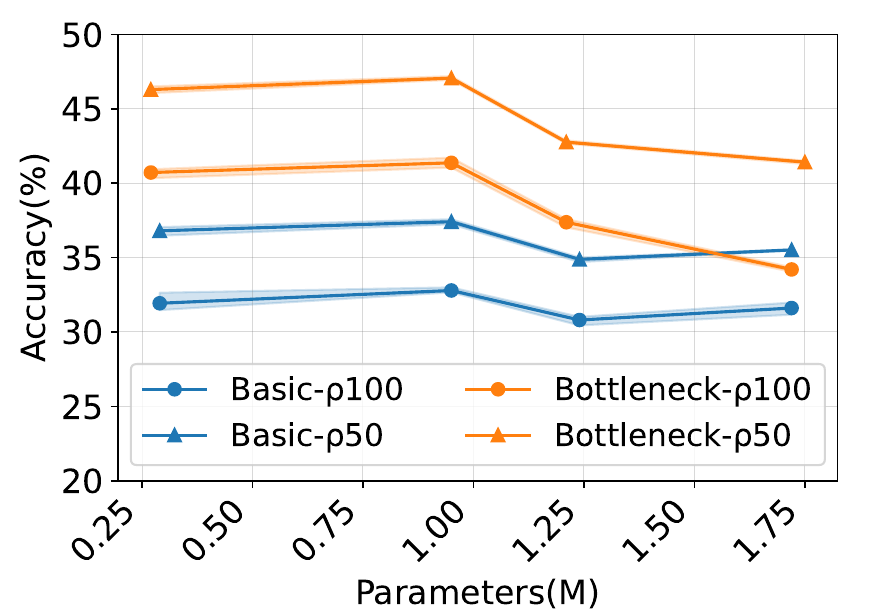}
        \caption{CIFAR-100-LT}
        \label{fig:basic_bottle_100}
    \end{subfigure}
    \caption{Comparison between ``Basic'' and ``Bottleneck'', and $\rho$ denotes the imbalance ratio.}
    \label{topo}
\end{figure}

\subsection{Topology}
We explore the performance of basic and bottleneck topology firstly, which are proposed in \cite{He2015DeepRL}. While the bottleneck has been proven to perform better on balanced datasets, many existing studies in the LT community default to using the basic topology in the backbone network~\cite{Cao2019LearningID,zhou2023ccsam}. We evaluate the network with basic and bottleneck topology on the LT dataset. The experimental results in Fig~\ref{topo} demonstrate that the bottleneck topology outperforms the basic, which aligns with the balanced context and suggests that the bottleneck may be a better choice in the search space.

\begin{figure}[tbp]
    \centering
    \begin{subfigure}[b]{0.23\textwidth}
        \centering
        \includegraphics[width=\textwidth]{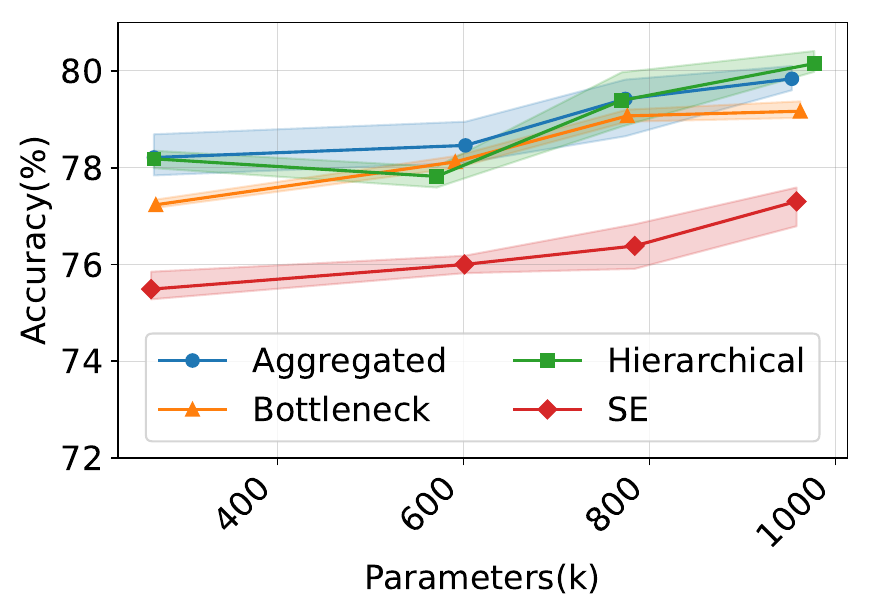}
        \caption{CIFAR-10-LT-$\rho$50}
        \label{fig:cov_50}
    \end{subfigure}
    \hfill
    \begin{subfigure}[b]{0.23\textwidth}
        \centering
        \includegraphics[width=\textwidth]{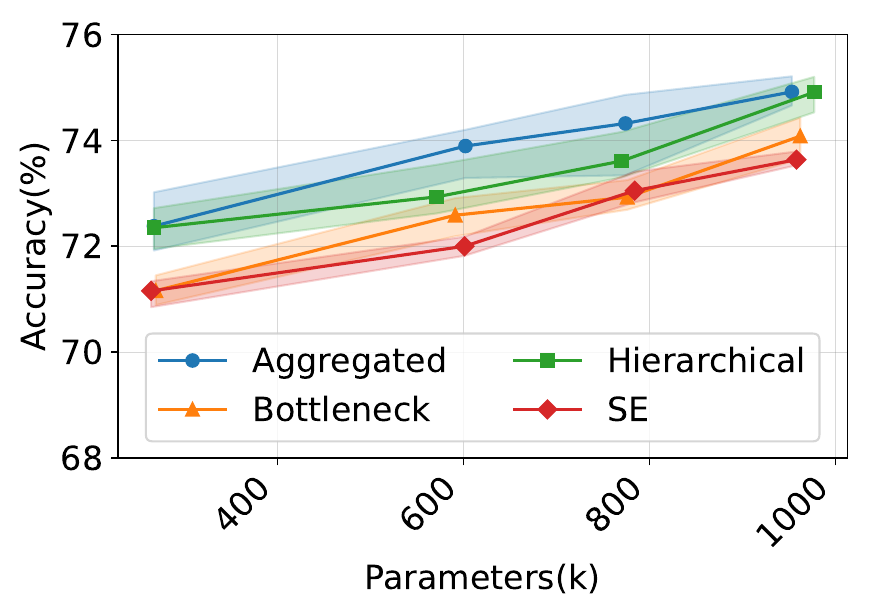}
        \caption{CIFAR-10-LT-$\rho$100}
        \label{fig:cov_100}
    \end{subfigure}

    \caption{Performance of different convolutions on CIFAR-10-LT with $\rho=\{50, 100\}$. 
    }
    \label{fig:conv}
\end{figure}

\subsection{Specific Convolution Design}
The specific design of convolutions primarily determines the network's ability to feature representation. In addition to the basic convolution operation (where we use the bottleneck topology), we further explore aggregate convolution~\cite{Xie2016AggregatedRT}, hierarchical convolution~\cite{Gao2019Res2NetAN}, and squeeze and excitation (SE) convolution~\cite{Hu2017SqueezeandExcitationN}, which are all proven effective on the balanced dataset.
As shown in Fig.~\ref{fig:explore_com}, the aggregate architecture extends the single path of conventional convolution into multiple paths, while the hierarchical architecture incorporates several sub-paths within the convolution to propagate feature information. The SE module dynamically adjusts the weight of each channel, enhancing the model's ability to focus on important features. These optimized convolutions have been proven efficient on balanced datasets, and we observe their performance on LT datasets. 
Fig.~\ref{fig:conv} shows that both the aggregated and hierarchical architectures enhance performance on LT data. However, the SE module leads to a performance drop, unlike its behavior on balanced datasets.

\subsection{Placement of Activation}
\begin{figure}
    \centering
        \begin{subfigure}[b]{0.23\textwidth}
        \centering
        \includegraphics[width=\textwidth]{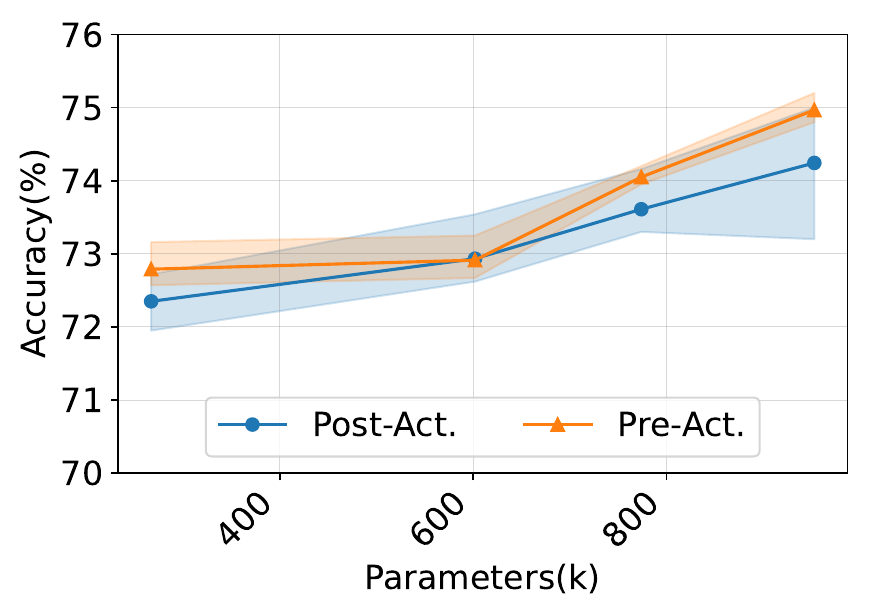}
        \caption{Aggregated Convolution}
        \label{fig:pre_post_agg}
    \end{subfigure}    
    \hfill
        \begin{subfigure}[b]{0.23\textwidth}
        \centering
        \includegraphics[width=\textwidth]{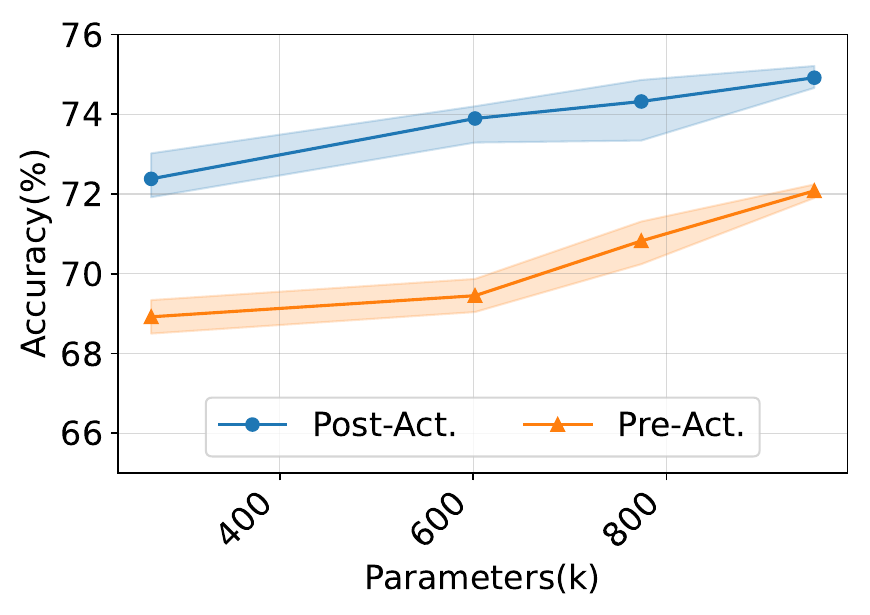}
        \caption{Hierarchical Convolution}
        \label{fig:pre_post_hier}
    \end{subfigure}
    \caption{Investigating the impact of activation function placement on the performance of aggregated convolution and hierarchical convolution.}
    \label{fig:poa}
\end{figure}

According to~\cite{he2016identity}, the placement of the activation function also impacts model performance. We conduct experiments to explore the optimal position of the activation function under different convolution types. Since the SE module has been shown to underperform compared to basic convolution, we focus only on the aggregated and hierarchical convolutions. Specifically, we examine two methods: pre-activation and post-activation. The results, presented in Fig.~\ref{fig:poa}, show that pre-activation benefits aggregation convolution, while post-activation is more effective for hierarchical convolution.

\subsection{Activation}
We study three different activation functions: ReLU~\cite{Glorot2011DeepSR}, ELU~\cite{clevert2016fast}, and Mish~\cite{Misra2019MishAS}. ReLU is highly efficient and widely used in many deep-learning models. ELU (Exponential Linear Unit) improves upon ReLU by allowing negative values, which helps mitigate the issue of dying neurons and can accelerate convergence. Mish, a newer activation function, combines the benefits of both ReLU and ELU, offering smoother gradients and improving model performance, particularly in deeper networks. We evaluate the performance of these activation functions in the context of LT data to determine which one provides the best results. Fig.~\ref{fig:act} reveals that ReLU remains the superior choice. 

\subsection{Normalization}
\begin{figure}[t]
    \centering
    \begin{subfigure}[b]{0.23\textwidth}
        \centering
        \includegraphics[width=\textwidth]{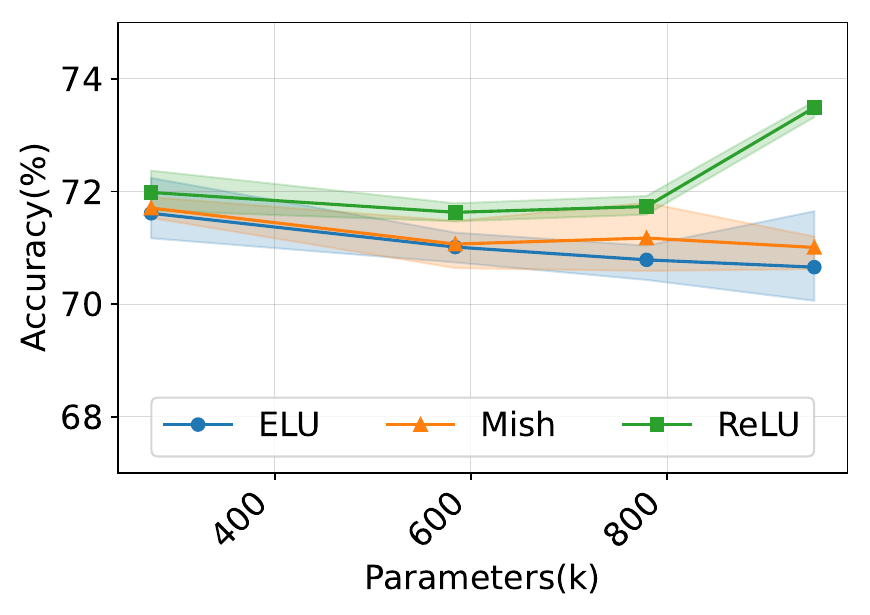}
        \caption{Activation}
        \label{fig:act}
    \end{subfigure}
    \hfill    
    \begin{subfigure}[b]{0.23\textwidth}
        \centering
        \includegraphics[width=\textwidth]{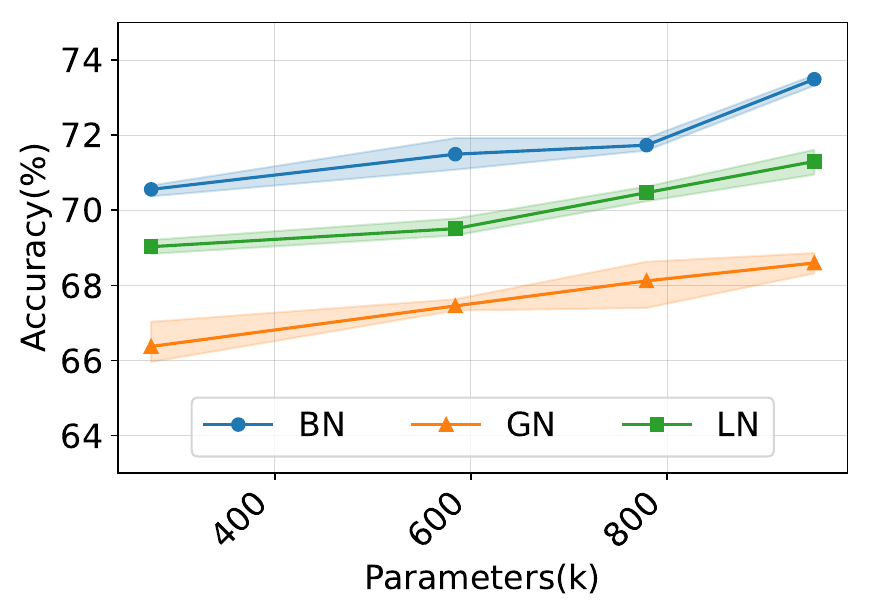}
        \caption{Normalization}
        \label{fig:norm}
    \end{subfigure}
    \label{fig:act_norm_app}
    \caption{(a) Performance of different activation functions on CIFAR-10-LT. (b) Impact of different normalization methods on model performance.}
\end{figure}

We investigate three normalization methods: BatchNorm~\cite{Ioffe2015BatchNA}, GroupNorm~\cite{Wu2018GroupN}, and LayerNorm~\cite{Ba2016LayerN}, to explore their performance on long-tailed datasets. As shown in Fig.~\ref{fig:norm}, batch normalization achieves the best performance under comparable model sizes. This suggests that BatchNorm’s ability to stabilize training by normalizing feature distributions per batch is particularly effective in mitigating the challenges posed by class imbalance in LT data. Therefore, we employ BatchNorm when constructing convolutional operations.

\begin{figure}[tbp]
    \centering
    \begin{subfigure}[b]{0.22\textwidth}
        \centering
        \includegraphics[width=\textwidth]{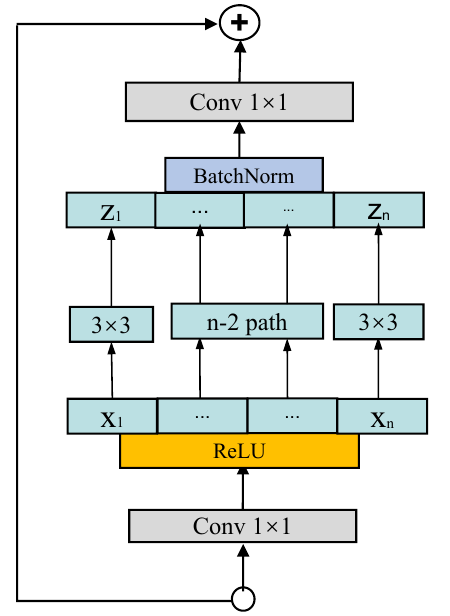}
        \caption{LT-AggConv}
        \label{fig:Agg}
    \end{subfigure}
    \hfill
    \begin{subfigure}[b]{0.22\textwidth}
        \centering
        \includegraphics[width=\textwidth]{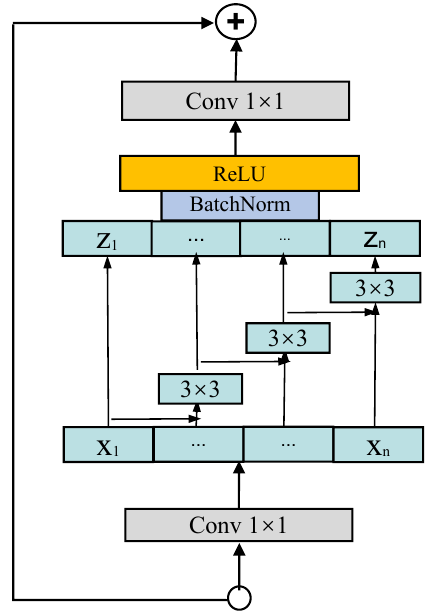}
        \caption{LT-HierConv}
        \label{fig:HierConv}
    \end{subfigure}
    \caption{Architecture of proposed LT-AggConv and LT-HierConv operators.}
    \label{fig:proposed_conv}
\end{figure}

\subsection{Operations Design}\label{Operations-Design}
Based on these observations, we design two convolutional operations specifically tailored for LT issues: LT-AggConv and LT-HierConv. The design details of the two novel convolutional operations are depicted in Fig.~\ref{fig:proposed_conv}. LT-AggConv employs an aggregated convolution with a pre-activation approach, which uses multiple parallel convolutional operations and combines their outputs. This design benefits LT scenarios by providing more refined feature representations for rare classes. LT-HierConv, on the other hand, uses hierarchical convolution with a post-activation approach, allowing the model to adaptively emphasize relevant features at various hierarchical levels, thereby enhancing classification accuracy. Both operations utilize the ReLU activation function and BatchNorm normalization based on a bottleneck topology. 

\begin{table}[tbp]
\centering
\caption{Performance comparison of LT-Res with existing ResNet series networks on long-tailed data.}
\resizebox{0.95\linewidth}{!}{
\begin{tabular}{cc|ccc}
\hline   
&  & \multicolumn{3}{c}{CIFAR-100-LT} \\
\hline
\multicolumn{1}{c|}{Backbone} & $\#$P(M) & $\rho$=200 & $\rho$=100 & $\rho$=50 \\ 
\hline
\multicolumn{1}{c|}{ResNet} & 0.46  & 34.75$\scriptstyle{\pm0.25}$ & 38.38$\scriptstyle{\pm0.18}$ & 43.72$\scriptstyle{\pm0.17}$ \\
\multicolumn{1}{c|}{ResNeXt} & 0.45 & 36.40$\scriptstyle{\pm0.26}$ & 41.14$\scriptstyle{\pm0.15}$ & 46.52$\scriptstyle{\pm0.28}$ \\
\multicolumn{1}{c|}{Res2Net} & 0.45 & 35.84$\scriptstyle{\pm0.33}$ & 40.18$\scriptstyle{\pm0.22}$ & 46.33$\scriptstyle{\pm0.21}$ \\
\multicolumn{1}{c|}{LT-Res} & 0.45 & \textbf{39.71}$\scriptstyle{\pm0.14}$ & \textbf{41.65}$\scriptstyle{\pm0.21}$ & \textbf{47.04}$\scriptstyle{\pm0.27}$ \\ 
\hline
\end{tabular}
}
\label{res-result}
\end{table}

To demonstrate the superiority of the proposed approach, we craft an LT-friendly residual network based on the ResNet prototype, named LT-Res. Specifically, we replace the convolution operations in ResNet with LT-AggConv and LT-HierConv and construct the network according to the base architecture outlined in Section~\ref{pre}. To highlight the effectiveness of LT-Res, we compare it against ResNet, ResNeXt, and Res2Net. As shown in Table~\ref{res-result}, performance gains are achieved through architecture optimization alone.

\section{A NAS Method for LT: LT-DARTS}

\begin{figure*}[t]
    \centering
    \begin{subfigure}[b]{0.23\textwidth}
        \centering
        \raisebox{0.15\height}{\includegraphics[width=\textwidth]{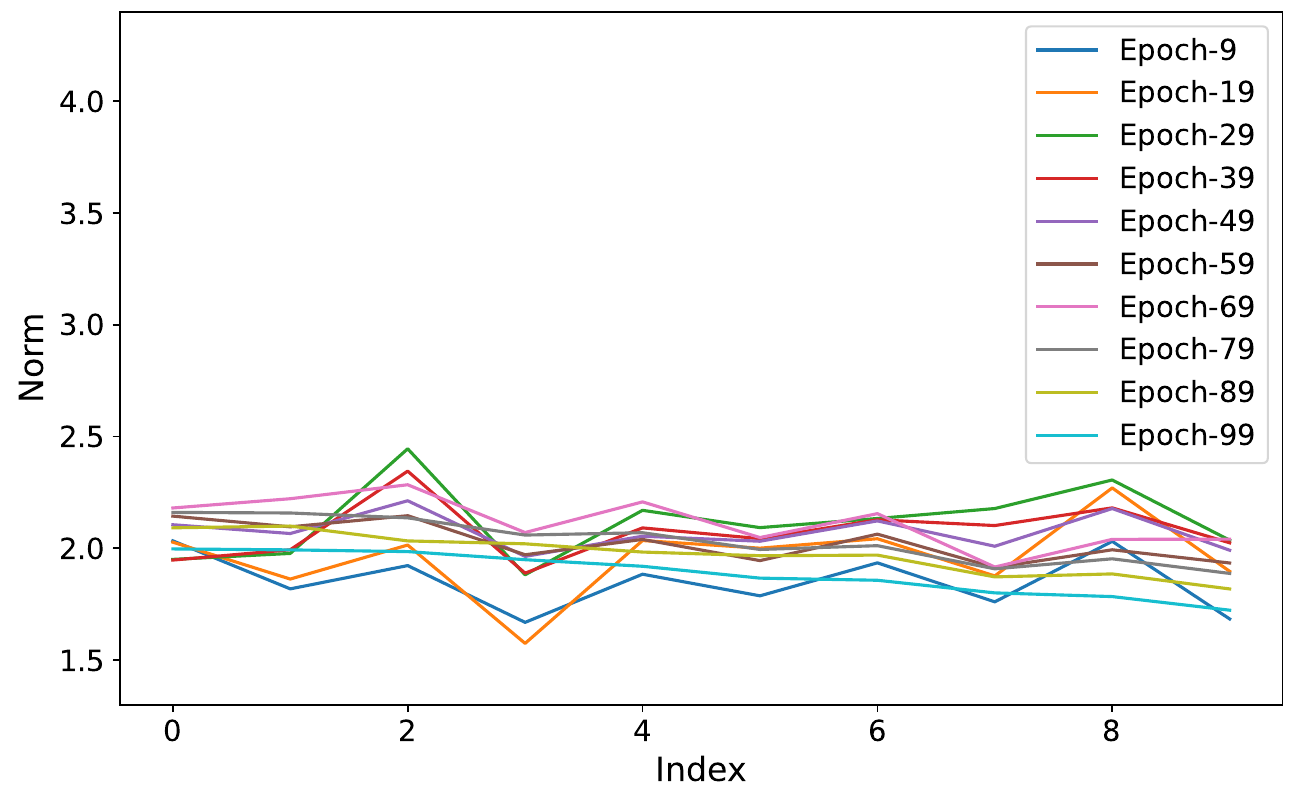}}
        \caption{weight norms on balanced}
        \label{fig:obv1}
    \end{subfigure}
    \hfill
    \begin{subfigure}[b]{0.23\textwidth}
        \raisebox{0.15\height}{\includegraphics[width=\textwidth]{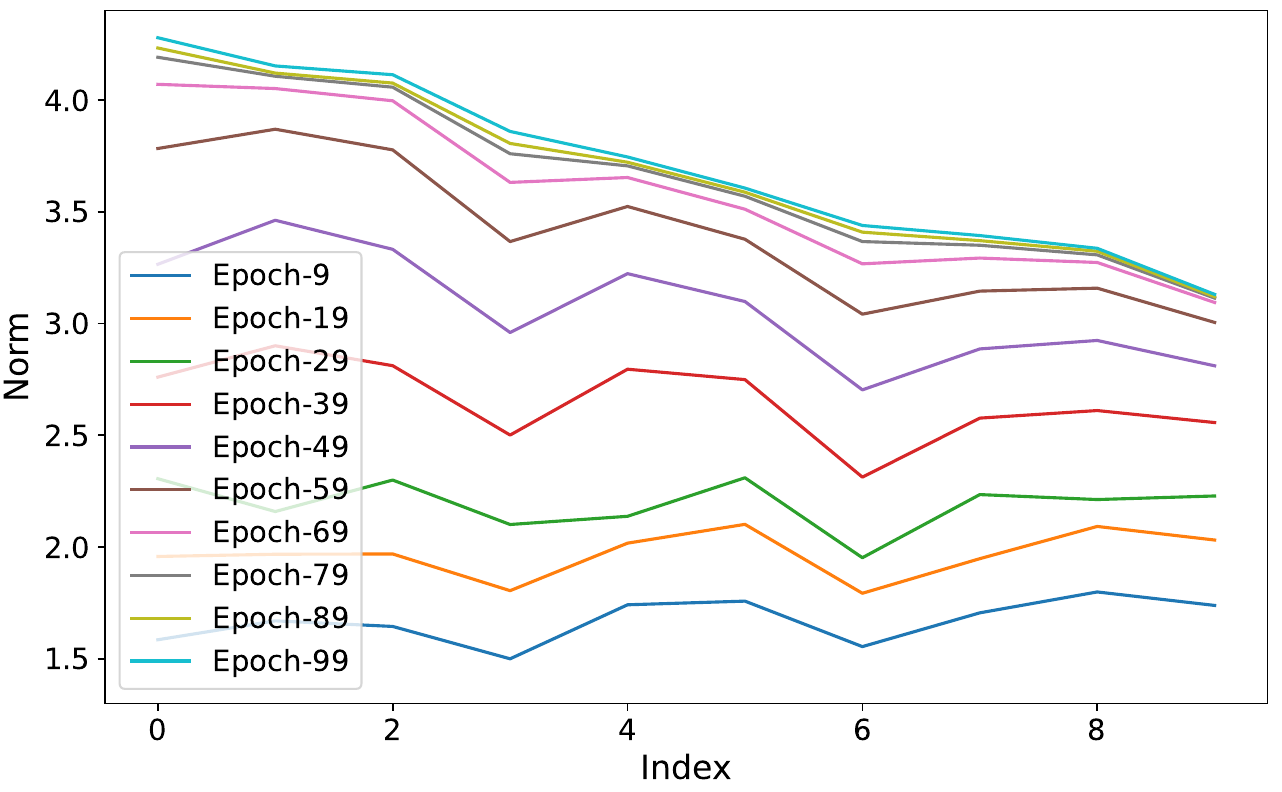}}
        \caption{weight norms on imbalanced}
        \label{fig:obv2}
    \end{subfigure}
    \hfill
    \begin{subfigure}[b]{0.23\textwidth}
        \centering
        \includegraphics[width=\textwidth]{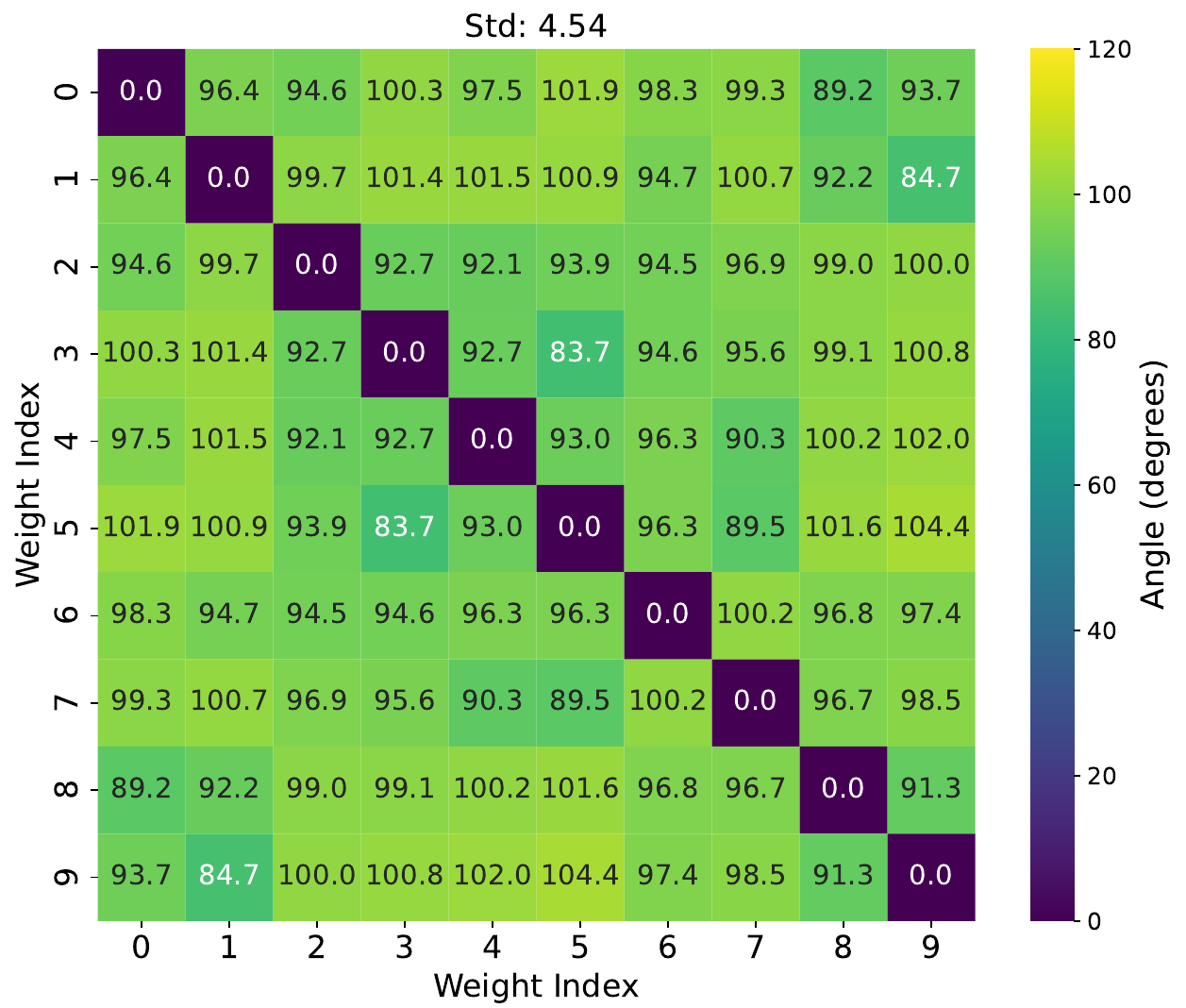}
        \caption{weight angles on balanced}
        \label{fig:obv3}
    \end{subfigure}
    \hfill
    \begin{subfigure}[b]{0.23\textwidth}
        \centering
        \includegraphics[width=\textwidth]{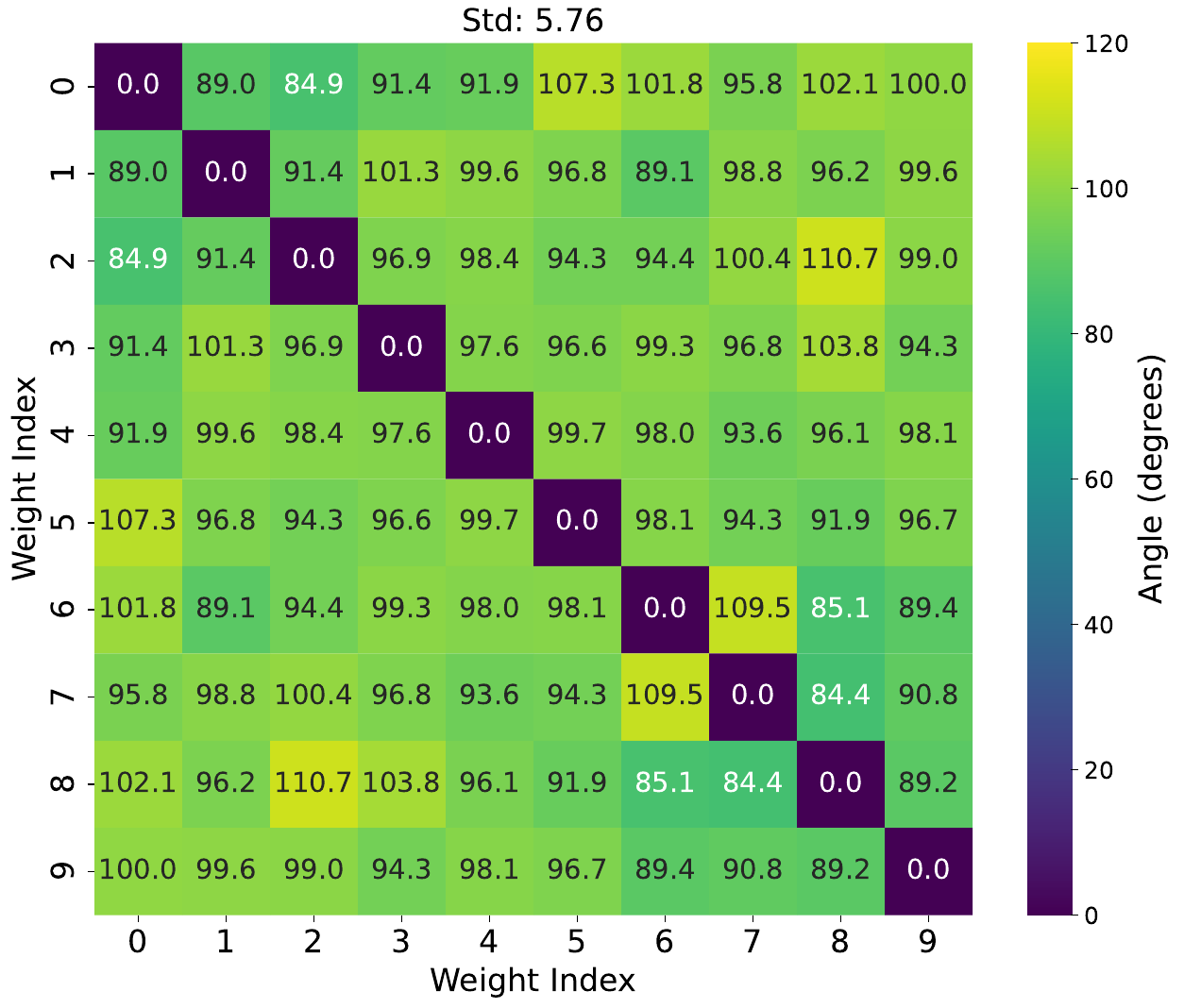}
        \caption{weight angles on imbalanced}
        \label{fig:obv4}
    \end{subfigure}
    \caption{Classifier weight biased phenomena. (a) and (b) show weight norms at different search epochs, highlighting shifts in weight norms with long-tailed data. (c) and (d) reveal that the angle distribution between weight vectors is more uneven under long-tailed distributions. Best viewed in color.}
    \label{fig:obv}
\end{figure*}

To design networks better suited for LT, we propose LT-DARTS, a straightforward yet effective NAS method optimized for LT challenges. Although our proposed LT convolutions demonstrate effectiveness, previous study highlights that the interconnections among architectural components are also crucial~\cite{Zhang2015ASA}. Exhaustively exploring these connectivity patterns can be resource-intensive; thus, we leverage NAS to efficiently address this complexity. Specifically, we enhance DARTS~\cite{Liu2018DARTSDA}, a gradient-based NAS approach, with targeted improvements in both the search space and strategy, leading to the development of LT-DARTS.
\subsection{Preliminaries}
DARTS constructs the neural network by stacking cells with the same architecture, where a cell is defined as a directed acyclic graph (DAG). The goal of search process is to identify the operations (edges) utilized for connecting diverse features (nodes). To this end, DARTS makes use of continuous relaxation in the following form:
\begin{equation}
    \bar{o}(x) = \sum_{o\in \mathcal{O} }\frac{\mathrm{exp}(\alpha _o )}{\sum_{o^{'}\in\mathcal{O} }\mathrm{exp}(\alpha _{o^{'}} )} o(x) 
\end{equation}
where $\mathcal{O}$ is the search space and $o(\cdot)$ is the different operations selected from $\mathcal{O}$. $\bar{o}(\cdot)$ is the mixed operation, and it will be replaced by the $o(\cdot)=\mathop{\mathrm{argmax}}_{o\in \mathcal{O}}{\alpha _{o}}$ at the end of the search. $\alpha$ is a vector of dimension $\left|\mathcal{O}\right|$, named architecture parameter, denotes the operation mixing  weights. It can be optimized through the following steps:
\begin{equation}
\label{darts_loss}
       \min_{\alpha }  \mathcal{L}_{val}(w^*(\alpha ), \alpha ) 
\end{equation}
\begin{equation}
\label{darts_loss_train}
    s.t.  \quad  w^*(\alpha )=\mathop{\mathrm{argmin}}\limits_{w}{\mathcal{L}_{train}(w,a)}
\end{equation}
where $w$ denotes the network parameters and $w^*(\alpha )$ is the optimal network parameters $w$ under the current architecture $\alpha$. $\mathcal{L}_{train}$ and  $\mathcal{L}_{val}$ denote the training and validation losses, respectively. The search process is based on gradient optimization of $\alpha$ and $w$ to find the optimal cell architecture.

\subsection{Search Space towards LT-DARTS}
We construct an LT-friendly search space, denoted as  $\mathcal{O}$. Following the previous approaches~\cite{Xu2019PCDARTSPC, Ye2022betaDARTSBR, Ye2023DARTSBR}, we include operations like $3 \times 3$ max pooling, $3 \times 3$ average pooling, residual connections, and zero (representing no connection) within the search space. Uniquely, based on our observations, we additionally introduce LT-AggConv and LT-HierConv as candidate convolution operations, each available with kernel sizes of $ \left\{3 \times 3, 5 \times 5, 7 \times 7 \right\}$. Other configuration settings follow those established by~\cite{Liu2018DARTSDA}. Each cell in our settings consists of 7 nodes: two input nodes, four intermediate nodes, and one output node. The output node serves as an aggregation of all other nodes within the cell. We form the network by stacking multiple cells, setting the first and second nodes of cell \( k \) equal to the outputs of cells \( k-2 \) and \( k-1 \), respectively. To enhance computational efficiency, we insert a reduction cell at $2/3$ of the network depth, where all operations adjacent to the input nodes use a stride of $2$.


\subsection{Search Strategy towards LT-DARTS}
To the best of our knowledge, no previous work has investigated the performance of DARTS on LT. In light of this, we conduct a systematic study and find that directly applying the vanilla DARTS to LT leads to suboptimal performance, as the linear classifier introduces bias during the search process. To overcome this issue, we introduce a new classifier with frozen parameters, updating only the backbone weights during the search process. This approach mitigates the bias introduced by the classifier and facilitates the discovery of more effective architectures.

\subsubsection{Observation: A Biased Search}
Previous studies have identified classifier bias in LT model training~\cite{Kang2019DecouplingRA, Nam2023DecoupledTF}. Similarly, we observe a gradual shift in the classifier during the network search process. As shown in Fig.~\ref{fig:obv}, on a balanced dataset, inter-class weight norms remain similar and show little change during the search process. In contrast, the differences in inter-class norms increase as the search progresses on LT data. Additionally, in the final epoch, the angles between weight vectors become more unevenly distributed on LT data compared to balanced data. Such biased norms and angles lead to a biased search, where head classes dominate the loss gradients, causing the learned architectures to overfit features that disproport.

\subsubsection{Balanced Fixed Classifier for Search Decoupling}
To address the bias search process, we introduce a Balanced Fixed Classifier (BFC) into the NAS process, a static, class-agnostic linear layer designed to decouple the architecture search from class frequency bias. This design draws inspiration from the phenomenon of neural collapse~\cite{Yang2022InducingNC}, where well-trained networks tend to converge toward a configuration where feature means and classifier weights form an Equiangular Tight Frame (ETF). Rather than treating this as a post-hoc outcome, we adopt this geometric structure as a prior and impose it from the start of the search.

Formally, the classifier weights $\mathbf{W} \in \mathbb{R}^{d \times C}$ are constructed to have equal norm and mutual angles, forming a centered ETF:
\begin{equation}
\label{eq:etf}
\mathbf{W} = \sqrt{E_W\frac{C}{C-1}}\mathbf{U}(\mathbf{I}-\frac{1}{C}\mathbf{1}\mathbf{1}^{\mathrm{T}})
\end{equation}
where $\mathbf{U} \in \mathbb{R}^{d \times C}$ is an orthonormal basis satisfying $\mathbf{U}^{\mathrm{T}}\mathbf{U} = \mathbf{I}$, and $E_W$ controls the weight scale. The ETF structure ensures that each class occupies an equal and uniformly distributed position in the weight space, regardless of its sample count.

Crucially, we freeze the classifier throughout the architecture search, updating only the backbone parameters. This prevents class imbalance from distorting the gradient flow during optimization, effectively neutralizing the dominance of head-class signals. As a result, the search process becomes more balanced and fair, encouraging the discovery of architectures that are not just optimized for frequent classes, but are also robust to tail-class sparsity.

\subsection{LT-DARTS Algorithm Overview}
We formalize the overall execution of the proposed LT-DARTS. Compared with vanilla DARTS, LT-DARTS incorporates two key modifications: (1) an LT-aware search space that includes LT-AggConv and LT-HierConv operators, and (2) a search strategy with a BFC to mitigate bias during optimization. The BFC adopts an ETF structure and remains frozen throughout the search, ensuring fairer gradient updates across head and tail classes. The search alternates between updating network weights $w$ using the training set $\mathcal{D}_{\text{train}}$ and updating architecture parameters $\alpha$ using the validation set $\mathcal{D}_{\text{val}}$. After $T$ epochs, the final discrete architecture is obtained by selecting the strongest operation on each edge. The full procedure is described in Algorithm~1.

\begin{algorithm}[h]
\caption{LT-DARTS}
\KwIn{Training data $\mathcal{D}_{\text{train}}$, Validation data $\mathcal{D}_{\text{val}}$, Search space $\mathcal{O}$, Total epochs $T$, Network $f_{\alpha, w}$}
\KwOut{Final discrete architecture $\mathcal{A}$}

Initialize network weights $w$ and architecture parameters $\alpha$\;

Construct a balanced fixed classifier $\mathbf{W}_{\text{BFC}}$ as Eq.~\ref{eq:etf}

\For{$t = 1$ \KwTo $T$}{
    Sample mini-batches $\mathcal{B}_{\text{train}} \subset \mathcal{D}_{\text{train}}$, $\mathcal{B}_{\text{val}} \subset \mathcal{D}_{\text{val}}$\;

    \tcp{Update network weights $w$}
    Compute training loss: \\
    $\mathcal{L}_{\text{train}} = \text{CE}(f_{\alpha, w}(\mathcal{B}_{\text{train}}), \mathbf{W}_{\text{BFC}})$\;
    Update $w \leftarrow w - \eta_w \nabla_w \mathcal{L}_{\text{train}}$\;

    \tcp{Update architecture parameters $\alpha$}
    Compute validation loss: $\mathcal{L}_{\text{val}} = \text{CE}(f_{\alpha, w}(\mathcal{B}_{\text{val}}), \mathbf{W}_{\text{BFC}})$\;
    Update $\alpha \leftarrow \alpha - \eta_\alpha \nabla_\alpha \mathcal{L}_{\text{val}}$\;
}

\tcp{Discretize architecture}
\ForEach{edge $(i, j)$ in DAG cell}{
    Select $o^* = \arg\max_{o \in \mathcal{O}} \alpha^{(i, j)}_o$\;
    Add $o^*$ to architecture $\mathcal{A}$
}

\Return{$\mathcal{A}$}
\end{algorithm}

\section{Experiments}
In this section, we evaluate the performance of architectures discovered by the proposed LT-DARTS. We compare with state-of-the-art methods on various LT datasets and conduct ablation experiments to demonstrate the effectiveness of our approach.
\subsection{Dateset and Implementation}
We evaluate the model's performance across four datasets of varying scales: CIFAR10-LT, CIFAR100-LT, Places-LT, ImageNet-LT, and  iNaturalist-LT. Without loss of generality, we apply different imbalanced factors to CIFAR datasets. We compare our method with some state-of-the-art approaches including LDAM-DRW~\cite{Cao2019LearningID}, BBN~\cite{Zhou2019BBNBN}, Balanced Softmax~\cite{Ren2020BalancedMF}, MiSLAS~\cite{Zhong2020ImprovingCF}, LADE~\cite{Hong2020DisentanglingLD},  GCL~\cite{Li2021LongtailedVR}, RSG~\cite{Wang2021RSGAS}, ResLT~\cite{Cui2021ResLTRL}, DisAlign~\cite{Zhang2021DistributionAA}
and BGP~\cite{Wang2022BalancedGP}. 

\begin{table*}[htbp]
\centering
\caption{We compare our approach with some state-of-the-art manually designed architectures on the CIFAR10/100-LT dataset. Without loss of generality, all networks are trained to converge using the cross-entropy loss.}
\resizebox{0.77\linewidth}{!}{
\begin{tabular}{cc|ccc|ccc}
\hline   
& & \multicolumn{3}{c|}{CIFAR-10-LT} & \multicolumn{3}{c}{CIFAR-100-LT} \\
\hline
\multicolumn{1}{c|}{Backbone} & $\#$P(M) & $\rho$=200 & $\rho$=100 & $\rho$=50 & $\rho$=200 & $\rho$=100 & $\rho$=50 \\ 
\hline
\multicolumn{1}{c|}{ResNet~\cite{He2015DeepRL}} & 0.46 & 65.68$\scriptstyle{\pm0.25}$ & 70.70$\scriptstyle{\pm0.31}$ & 74.81$\scriptstyle{\pm0.17}$ & 34.84$\scriptstyle{\pm0.12}$ & 38.43$\scriptstyle{\pm0.09}$ & 43.90$\scriptstyle{\pm0.14}$ \\
\multicolumn{1}{c|}{ResNeXt~\cite{Xie2016AggregatedRT}} & 0.45 & 68.26$\scriptstyle{\pm0.34}$ & 74.14$\scriptstyle{\pm0.22}$ & 79.06$\scriptstyle{\pm0.27}$ & 36.44$\scriptstyle{\pm0.18}$ & 41.07$\scriptstyle{\pm0.11}$ & 46.61$\scriptstyle{\pm0.19}$ \\
\multicolumn{1}{c|}{Res2Net~\cite{Gao2019Res2NetAN}} & 0.45 & 67.92$\scriptstyle{\pm0.19}$ & 73.65$\scriptstyle{\pm0.14}$ & 78.23$\scriptstyle{\pm0.22}$ & 35.97$\scriptstyle{\pm0.32}$ & 40.28$\scriptstyle{\pm0.10}$ & 46.28$\scriptstyle{\pm0.21}$ \\ \hline
\multicolumn{1}{c|}{LT-Res} & 0.45 & \underline{69.35}$\scriptstyle{\pm0.25}$ & \underline{76.48}$\scriptstyle{\pm0.20}$ & \underline{81.44}$\scriptstyle{\pm0.29}$ & \underline{39.65}$\scriptstyle{\pm0.20}$ & \underline{41.55}$\scriptstyle{\pm0.24}$ & \underline{47.10}$\scriptstyle{\pm0.18}$ \\
\multicolumn{1}{c|}{LT-DARTS} & 0.46 & \textbf{70.23}$\scriptstyle{\pm0.18}$ & \textbf{77.52}$\scriptstyle{\pm0.14}$ & \textbf{82.45}$\scriptstyle{\pm0.33}$ & \textbf{41.23}$\scriptstyle{\pm0.22}$ & \textbf{42.31}$\scriptstyle{\pm0.15}$ & \textbf{47.58}$\scriptstyle{\pm0.21}$ \\ 
\hline
\end{tabular}
}
\label{res1}
\end{table*}

Our framework is implemented using PyTorch v1.13.0 and CUDA v11.7. For the specific details of LT-DARTS, we mostly followed the default settings of DARTS but made changes to certain parameters. Specifically, we use a smaller number of proxy layers (6 layers) to accelerate the search process and impose a larger weight decay $\mu$ = 6e-4 to enhance the generalizability of the model. For the comparison subjects lacking experimental results, we reproduce them following the default settings of their respective open-source codes.

\subsection{Main Results}
\subsubsection{Improvements Brought by Architectural Changes Only}
Firstly, we compare the performance of the best architecture discovered by LT-DARTS with ResNet, ResNeXt, and Res2Net on the CIFAR-10/100-LT datasets. It should be noted that Vision Transformer (ViT)\cite{Dosovitskiy2020AnII}, which typically involve much larger parameter counts, have not been widely adopted as baselines in previous LT learning research. Therefore, we do not include them in our comparative analysis. Additionally, to demonstrate the comprehensive exploratory capability of NAS, we also include LT-Res in our comparisons. The experimental results are shown in Table~\ref{res1}. It is evident that the architectures we proposed exhibit heightened performance within a similar model size. Specifically, both LT-Res and LT-DARTS yielded superior results. Due to the more thorough exploration conducted by LT-DARTS, it consistently outperformed in all scenarios. The architecture discovered by LT-DARTS shows a 4\% to 8\% improvement over the basic ResNet baseline. Even when compared to more advanced architectures like ResNeXt and Res2Net, it still achieves at least a 1\% accuracy advantage. Experiments across different datasets and imbalanced factors further validate the wide applicability of LT-DARTS.

\subsubsection{Achieving SOTA Performance by Combining Existing Methods}
Additionally, we compare our method with state-of-the-art (SOTA) methods in the LT community. Our approach aims to find an optimal architecture, which is orthogonal to most existing solutions, making it easy to combine with various methods for further performance enhancement. Table~\ref{tb:res_cifar} presents our experimental results. In each block of this table (except the last), the first row represents existing SOTA methods with a manually designed ResNet backbone, which is also the baseline. The second row shows SOTA methods combined with vanilla DARTS, and the third row features SOTA methods combined with our approach. 
We can conclude that: (i) Simply replacing the ResNet architecture in the baseline with LT-DARTS (CE+Ours) achieves superior performance compared to the SOTA method (BBN). It is important to note that the network is trained using the cross-entropy loss and does not incorporate any advanced rebalancing techniques. (ii)  After enhancing the search space and strategy, LT-DARTS is able to identify architectures better suited to LT data, significantly outperforming the baseline. (iii) Moreover, the architecture discovered by LT-DARTS surpasses those of meticulously designed LT strategies, only with simple data augmentation (Mixup) and classifier retraining strategies (cRT).

\begin{table}[t]
\centering
\caption{Comparison with SOTA methods. No specific annotation indicates using ResNet as the backbone.}
\resizebox{0.87\linewidth}{!}{
\begin{tabular}{l|ll|ll}
\hline
               & \multicolumn{2}{c|}{CIFAR-10-LT}                   & \multicolumn{2}{c}{CIFAR-100-LT}                  \\ \hline
               & \multicolumn{1}{c}{200} & \multicolumn{1}{c|}{100} & \multicolumn{1}{c}{200} & \multicolumn{1}{c}{100} \\ \hline
CE             &  65.68      &70.70      &    34.84             &38.43 \\
CE+DARTS       &  62.31      &67.29      &    31.46             &36.02 \\
CE+Ours        &  70.23      &77.52      &    41.23             &42.31 \\ \hline      
CE+Mixup~\cite{Zhang2017mixupBE}       & 65.84                   & 72.96                    & 35.84                   & 40.01                   \\
CE+Mixup+DARTS & 61.32                   & 68.21                    & 31.13                   & 37.25                   \\
CE+Mixup+Ours  & 73.87                   & 78.86                    & 43.51                   & 44.77         \\ \hline
LDAM-DRW~\cite{Cao2019LearningID}       & 75.32                   & 77.03                    & 38.91                   & 42.04                   \\
LDAM-DRW+DARTS & 72.56                   & 75.34                    & 36.24                   & 39.11                \\
LDAM-DRW+Ours  & 77.26                   & 79.61                    &44.45                    &45.79               \\ \hline
Mixup+cRT~\cite{Kang2019DecouplingRA}      & 73.06                   & 79.15                    & 41.73                   & 45.12                   \\
Mixup+cRT+DARTS& 70.97                   & 76.45                    & 38.05                   & 42.39  \\
Mixup+cRT+Ours & \textbf{80.50}          & \textbf{82.91}           &\textbf{44.90}          & \textbf{49.47}   \\ \hline

BBN~\cite{Zhou2019BBNBN}            & 73.47                   & 79.82                    & 37.21                   & 42.56                   \\
BGP~\cite{Wang2022BalancedGP}            & -                        & -                       & 41.20                   & 45.20                   \\
GCL~\cite{Li2021LongtailedVR}            & 79.03                  & 82.68                     & 44.88                   & 48.71                   \\         \hline
Optimal        & 80.50         & 82.91            &44.90         & 49.47                        \\\hline
\end{tabular}
}
\label{tb:res_cifar}
\end{table}

\subsubsection{Class-wise Accuracy under Long-Tailed Distribution}
To further verify how the proposed method improves the performance, we conduct a detailed class-wise performance analysis on the CIFAR-10-LT dataset with $\rho = 100$. Following standard practice in long-tailed recognition research~\cite{zhou2023ccsam}, we divide the samples into three disjoint groups based on the number of training samples per class: head classes, medium classes, and tail classes. We evaluate and report the top-1 accuracy within each group for several representative baselines and our proposed methods.

As shown in Table~\ref{tb:class-wise}, LT-DARTS achieves consistent improvements across all three class splits. This indicates that the searched architecture enhances the overall feature representation quality, regardless of class frequency. More importantly, the performance gain is especially prominent in the tail classes. For example, LT-DARTS improves tail-class accuracy from 41.0\% (ResNet) to 66.0\%, yielding a relative improvement of over 60\%. While the head class accuracy also improves (from 72.5\% to 77.2\%), the margin is relatively smaller. This disproportionate improvement in tail-class performance suggests that our architectural search method is particularly effective at enhancing the discriminative capacity for under-represented classes, without sacrificing the performance on frequent classes.These findings validate our hypothesis that architecture-level adaptation serves as a powerful and orthogonal approach to long-tailed learning.

\begin{table}[th]
\centering
\caption{Class-wise accuracy breakdown on CIFAR-10-LT.}
\resizebox{0.87\linewidth}{!}{\begin{tabular}{lcccc}
\toprule
\textbf{Backbone} & \textbf{Head} & \textbf{Medium} & \textbf{Tail} & \textbf{Overall} \\
\midrule
ResNet & 72.5 & 58.0 & 41.0 & 70.70 \\
ResNeXt & 75.0 & 66.5 & 52.0 & 74.14 \\
DARTS & 74.2 & 66.0 & 55.0 & 73.65 \\
LT-ResNet (ours) & \underline{76.5} & \underline{72.0} & \underline{64.0} & \underline{76.48} \\
\textbf{LT-DARTS (ours)} & \textbf{77.2} & \textbf{73.5} & \textbf{66.0} & \textbf{77.25} \\
\bottomrule
\label{tb:class-wise}
\end{tabular}}
\end{table}

\subsubsection{Large-scale Datasets}
We also validate the effectiveness of our method on several large-scale long-tailed benchmarks, including Places-LT, ImageNet-LT, and iNaturalist-LT. As shown in Table~\ref{tb:res_combined}, LT-DARTS achieves competitive or superior top-1 accuracy on all datasets, despite having fewer parameters than the commonly used backbones. For example, on ImageNet-LT, LT-DARTS achieves 54.9\% accuracy with only 22M parameters, outperforming DisAlign (53.4\%) built on ResNeXt-50 with 23M parameters. Similarly, on Places-LT and iNaturalist-LT, our architecture achieves 40.5\% and 71.0\% respectively, with lower model complexity compared to ResNet-152 and ResNet-50 backbones.

\begin{table}[t]
\centering
\caption{Top-1 Accuracy on Places-LT, ImageNet-LT, and iNaturalist-LT with different backbones.}
\resizebox{0.95\linewidth}{!}{
\begin{tabular}{l|c|l|l|l}
\hline
Method & Dataset & Backbone & \#P(M) & Acc. \\ \hline
\multirow{10}{*}{\shortstack[l]{}} 
CE & \multirow{10}{*}{Places-LT} & ResNet-152 & 58.2 & 30.2 \\
Decouple-cRT~\cite{Kang2019DecouplingRA} &  & ResNet-152 & 58.2 & 36.7 \\
Balanced Softmax~\cite{Ren2020BalancedMF} &  & ResNet-152 & 58.2 & 38.6 \\
LADE~\cite{Hong2020DisentanglingLD} &  & ResNet-152 & 58.2 & 38.8 \\
DisAlign~\cite{Zhang2021DistributionAA} &  & ResNet-152 & 58.2 & 39.3 \\
RSG~\cite{Wang2021RSGAS} &  & ResNet-152 & 58.2 & 39.3 \\
MiSLAS~\cite{Zhong2020ImprovingCF} &  & ResNet-152 & 58.2 & 40.2 \\
GCL~\cite{Li2021LongtailedVR} &  & ResNet-152 & 58.2 & \textbf{40.6} \\
Mixup+cRT &  & ResNet-152 & 58.2 & 38.5 \\
Mixup+cRT &  & \textbf{LT-DARTS} & \textbf{56.4} & \underline{40.5} \\ \hline

\multirow{8}{*}{\shortstack[l]{}} 
CE & \multirow{8}{*}{ImageNet-LT} & ResNeXt-50 & 23.0 & 44.4 \\
Balanced Softmax~\cite{Ren2020BalancedMF} &  & ResNeXt-50 & 23.0 & 52.3 \\
LADE~\cite{Hong2020DisentanglingLD} &  & ResNeXt-50 & 23.0 & 52.3 \\
DisAlign~\cite{Zhang2021DistributionAA} &  & ResNeXt-50 & 23.0 & \underline{53.4} \\
RSG~\cite{Wang2021RSGAS} &  & ResNeXt-50 & 23.0 & 51.8 \\
ResLT~\cite{Cui2021ResLTRL} &  & ResNeXt-50 & 23.0 & 52.9 \\
Mixup+cRT &  & ResNeXt-50 & 23.0 & 52.3 \\
Mixup+cRT &  & \textbf{LT-DARTS} & \textbf{22.0} & \textbf{54.9} \\ \hline

\multirow{8}{*}{\shortstack[l]{}} 
CE & \multirow{8}{*}{iNaturalist-LT} & ResNet-50 & 25.6 & 61.7 \\
Balanced Softmax~\cite{Ren2020BalancedMF} &  & ResNet-50 & 25.6 & \underline{70.6} \\
LADE~\cite{Hong2020DisentanglingLD} &  & ResNet-50 & 25.6 & 70.0 \\
DisAlign~\cite{Zhang2021DistributionAA} &  & ResNet-50 & 25.6 & \underline{70.6} \\
RSG~\cite{Wang2021RSGAS} &  & ResNet-50 & 25.6 & 70.3 \\
ResLT~\cite{Cui2021ResLTRL} &  & ResNet-50 & 25.6 & 70.2 \\
Mixup+cRT &  & \textbf{LT-DARTS} & \textbf{24.8} & \textbf{71.0} \\ \hline

\end{tabular}
}
\label{tb:res_combined}
\end{table}

Notably, these gains are obtained with minimal additional effort: our method is combined with simple long-tailed training techniques such as Mixup and cRT, without requiring complicated training pipelines, additional modules, or extensive hyperparameter tuning. Besieds, it is important to emphasize that our goal is not to compete purely for SOTA performance. Instead, we focus on showcasing the potential of architecture design as a clean and principled approach that improves model performance across various long-tailed setting.

\subsection{Architecture Analysis of Searched Cells}

\begin{figure}[t]
    \centering
    \begin{subfigure}[b]{0.5\textwidth}
        \centering
        \includegraphics[width=\textwidth]{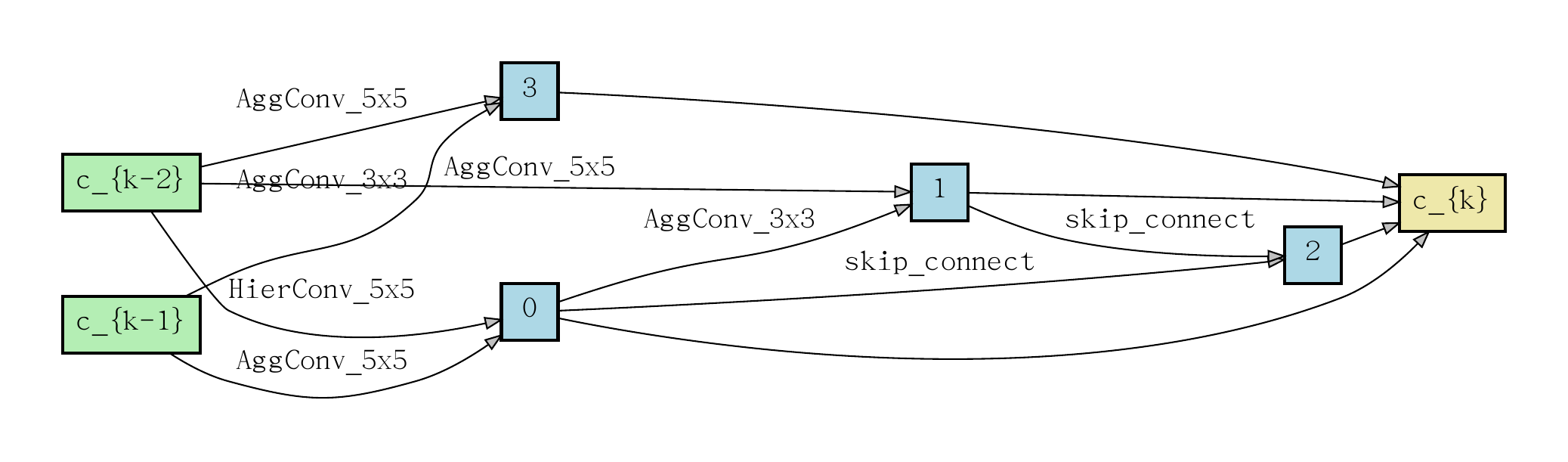}
        \caption{Normal cell of CIFAR-10-LT}
        \label{fig:norm_cell_cifar10}
    \end{subfigure} 
    \begin{subfigure}[b]{0.5\textwidth}
        \centering
        \includegraphics[width=\textwidth]{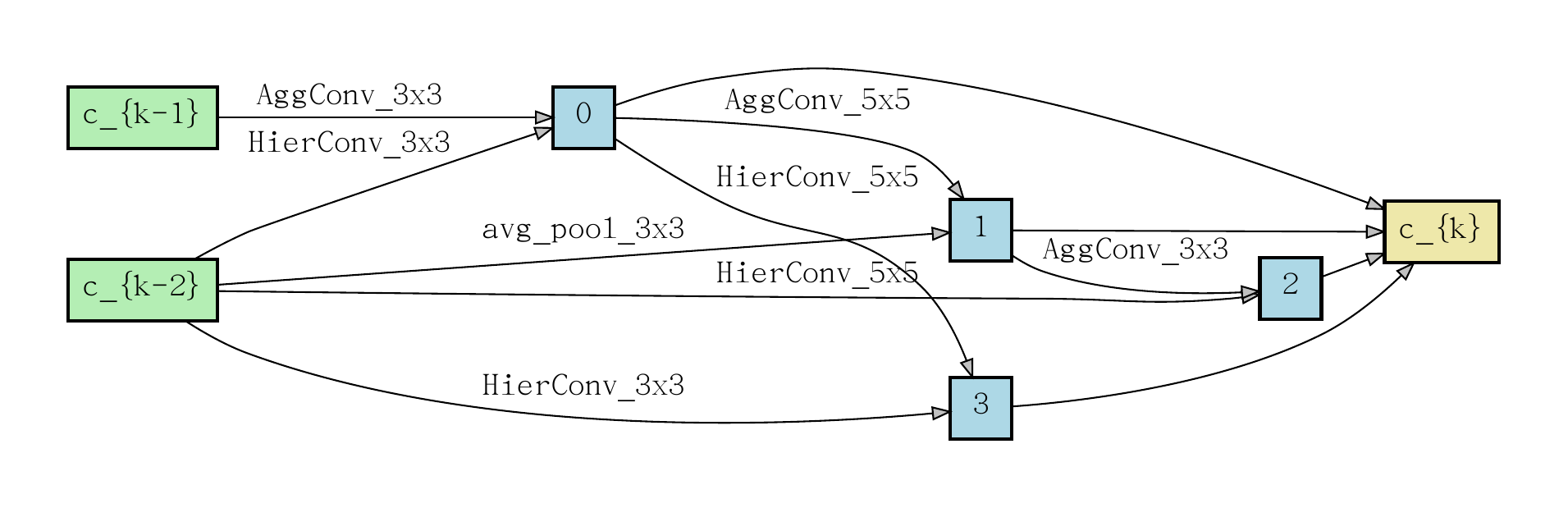}
        \caption{Reduction cell of CIFAR-10-LT}
        \label{fig:redu_cell_cifar10}
    \end{subfigure}
    \caption{Cells with optimal performance searched on the CIFAR-10-LT Dataset.}
    \label{fig:genotype_10}
\end{figure}

\begin{figure}[t]
    \centering
    \begin{subfigure}[b]{0.25\textwidth}
        \centering
        \includegraphics[width=\textwidth]{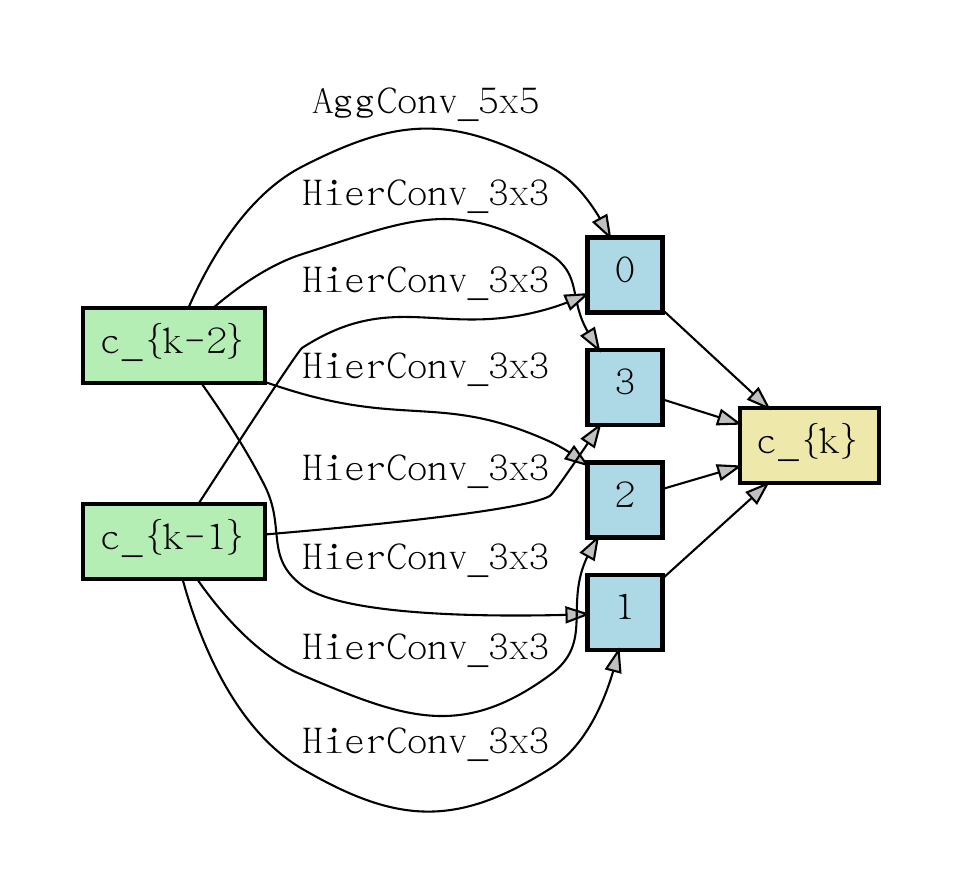}
        \caption{Normal cell of CIFAR-100-LT}
        \label{fig:norm_cell_cifar100}
    \end{subfigure} 
    \begin{subfigure}[b]{0.5\textwidth}
        \centering
        \includegraphics[width=\textwidth]{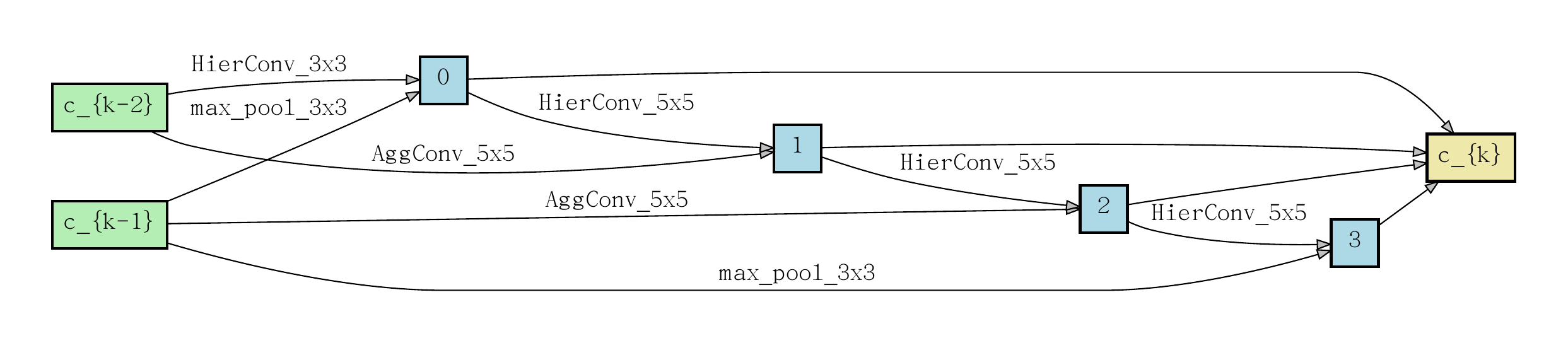}
        \caption{Reduction cell of CIFAR-100-LT}
        \label{fig:redu_cell_cifar100}
    \end{subfigure}
    \caption{Cells with optimal performance searched on the CIFAR-100-LT Dataset.}
    \label{fig:genotype_100}
\end{figure}

To better understand the structural patterns favored by LT-DARTS in LT scenarios, we analyze the architectures it discovers on CIFAR-10-LT and CIFAR-100-LT (see Fig.~\ref{fig:genotype_10} and Fig.~\ref{fig:genotype_100}). Across both datasets, the searched cells consistently prefer LT-HierConv and LT-AggConv. This suggests that the two custom operations are particularly effective for addressing class imbalance. In addition, the discovered architectures incorporate a mix of kernel sizes (e.g., 3×3 and 5×5), skip connections, and moderate pooling operations. These design choices facilitate multi-scale feature extraction and help stabilize training, which is crucial under LT conditions.

Compared to manually designed networks, LT-DARTS demonstrates two key advantages: more suitable operation and greater structural flexibility. The proposed LT-HierConv and LT-AggConv integrate empirically favorable components, allowing the network to extract the feature of both head and tailed classes well. Further, LT-DARTS benefits from a search-driven design that adaptively assembles heterogeneous operations, while the hand-designed networks constrained by fixed topologies and homogeneous block structures.

In conclusion, our approach enables the automatic design of flexible and adaptive neural architectures along both component and network levels, thereby improving generalization performance on tail classes without compromising accuracy on head classes. This marks a clear divergence from the fixed structures of traditional residual networks.

\begin{figure*}[t]
    \centering
    \begin{subfigure}[b]{0.30\textwidth}
        \centering
        \includegraphics[width=\textwidth]{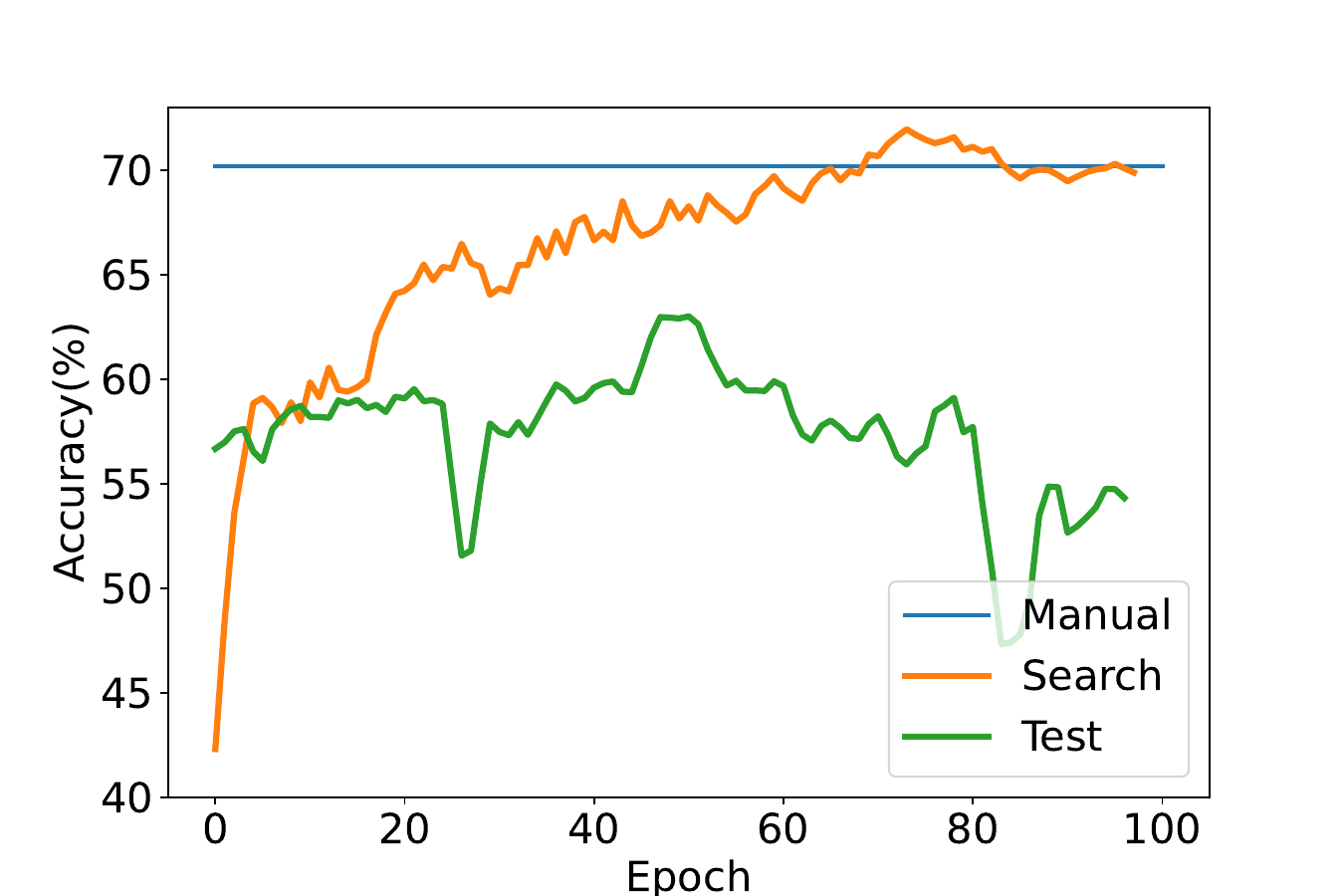}
        \caption{Vanilla}
        \label{fig:vanilla_test}
    \end{subfigure}    
        \begin{subfigure}[b]{0.30\textwidth}
        \centering
        \includegraphics[width=\textwidth]{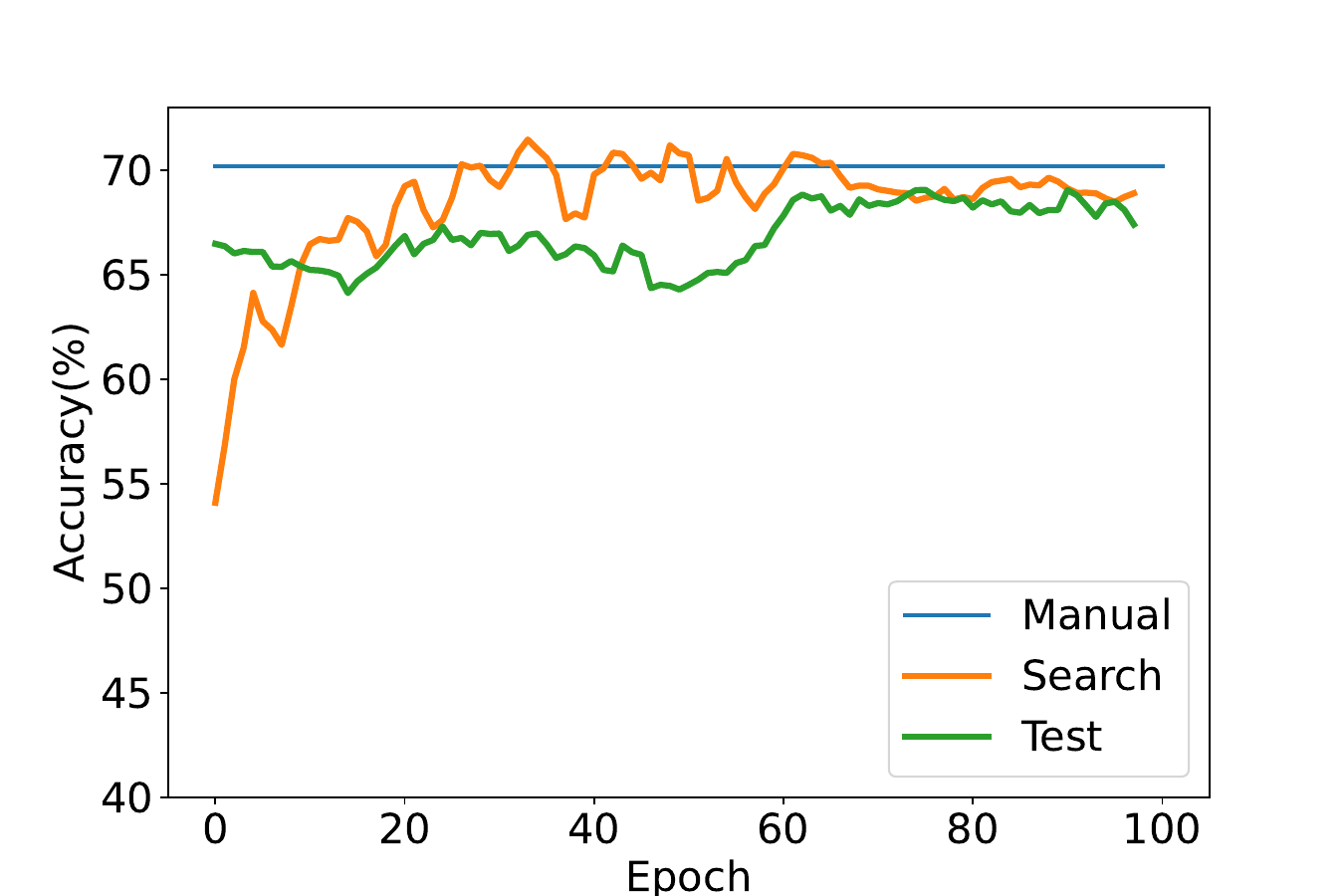}
        \caption{Balanced Fixed Classifier}
        \label{fig:etf_test}
    \end{subfigure}
    \begin{subfigure}[b]{0.29\textwidth}
        \centering
        \includegraphics[width=\textwidth]{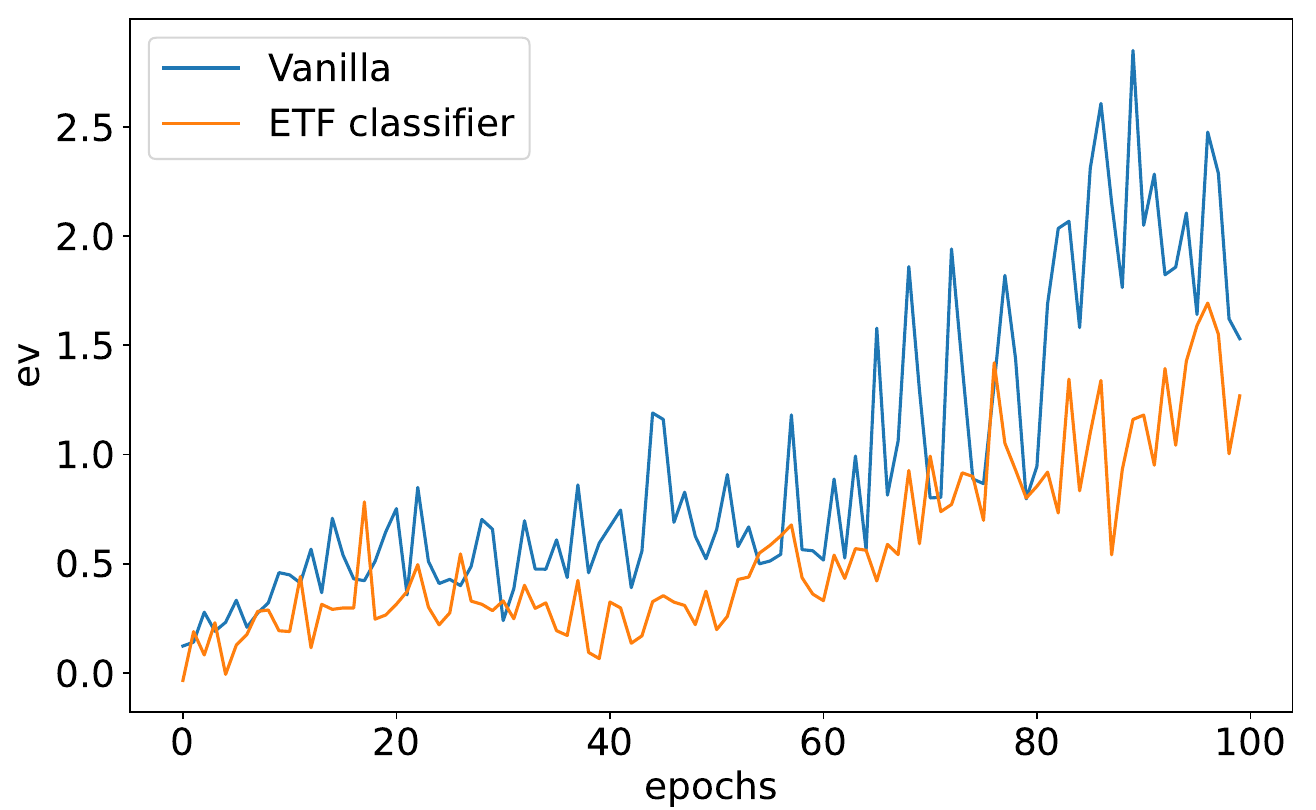}
        \caption{Eigenvalues}
        \label{fig:ev}
    \end{subfigure}
    \caption{(a) and (b) show the performance differences between trainable and balanced fixed classifiers. ``Search'' indicates the search process performance and ``test'' result is a snapshot of the corresponding network training 100 epochs from scratch. (c) report the trajectory of Hessian matrix eigenvalues during architecture search.}
    \label{fig:performance-cls}
\end{figure*}

\subsection{Benefits of the BFC}
\begin{figure}
    \begin{subfigure}[b]{0.23\textwidth}
        \centering
        \raisebox{0.15\height}{\includegraphics[width=\textwidth]{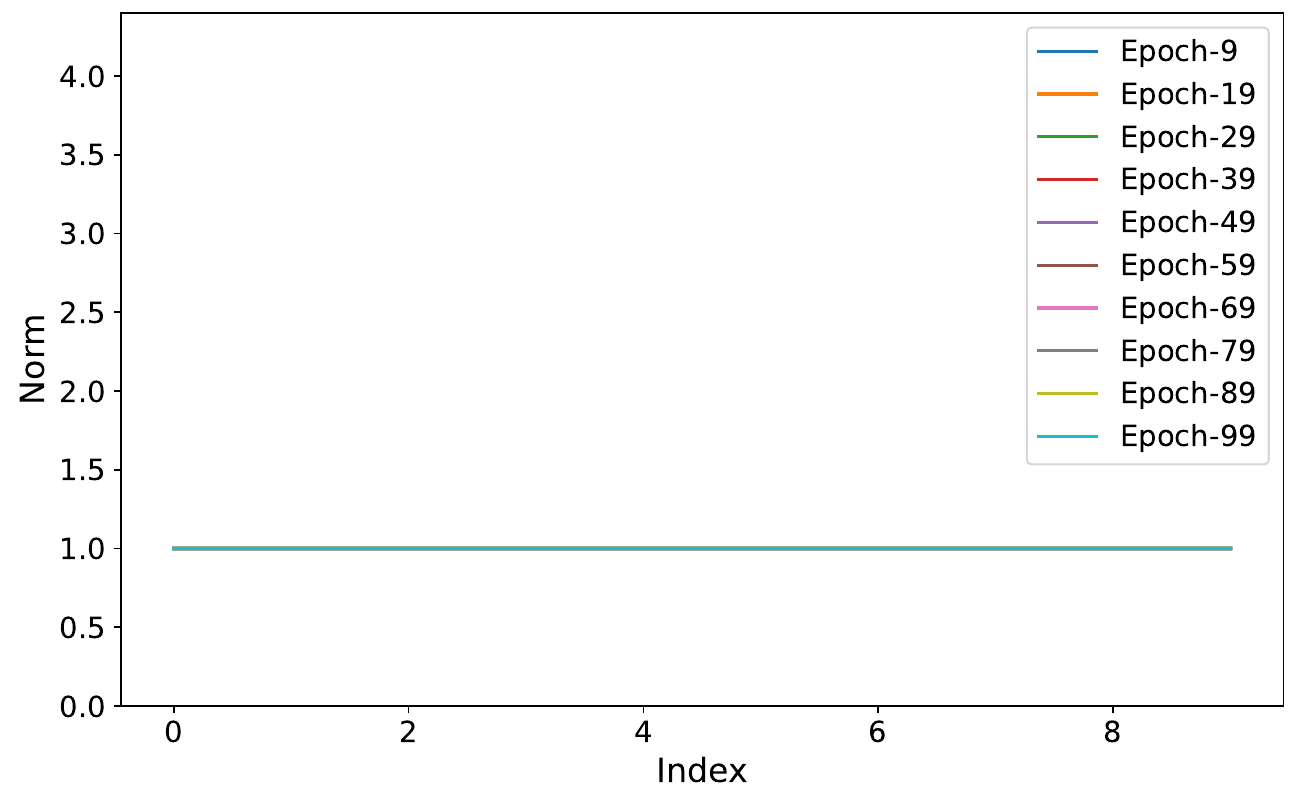}}
        \caption{weight norms}
        \label{fig:etf_norm}
    \end{subfigure}    
    \hfill
        \begin{subfigure}[b]{0.23\textwidth}
        \centering
        \includegraphics[width=\textwidth]{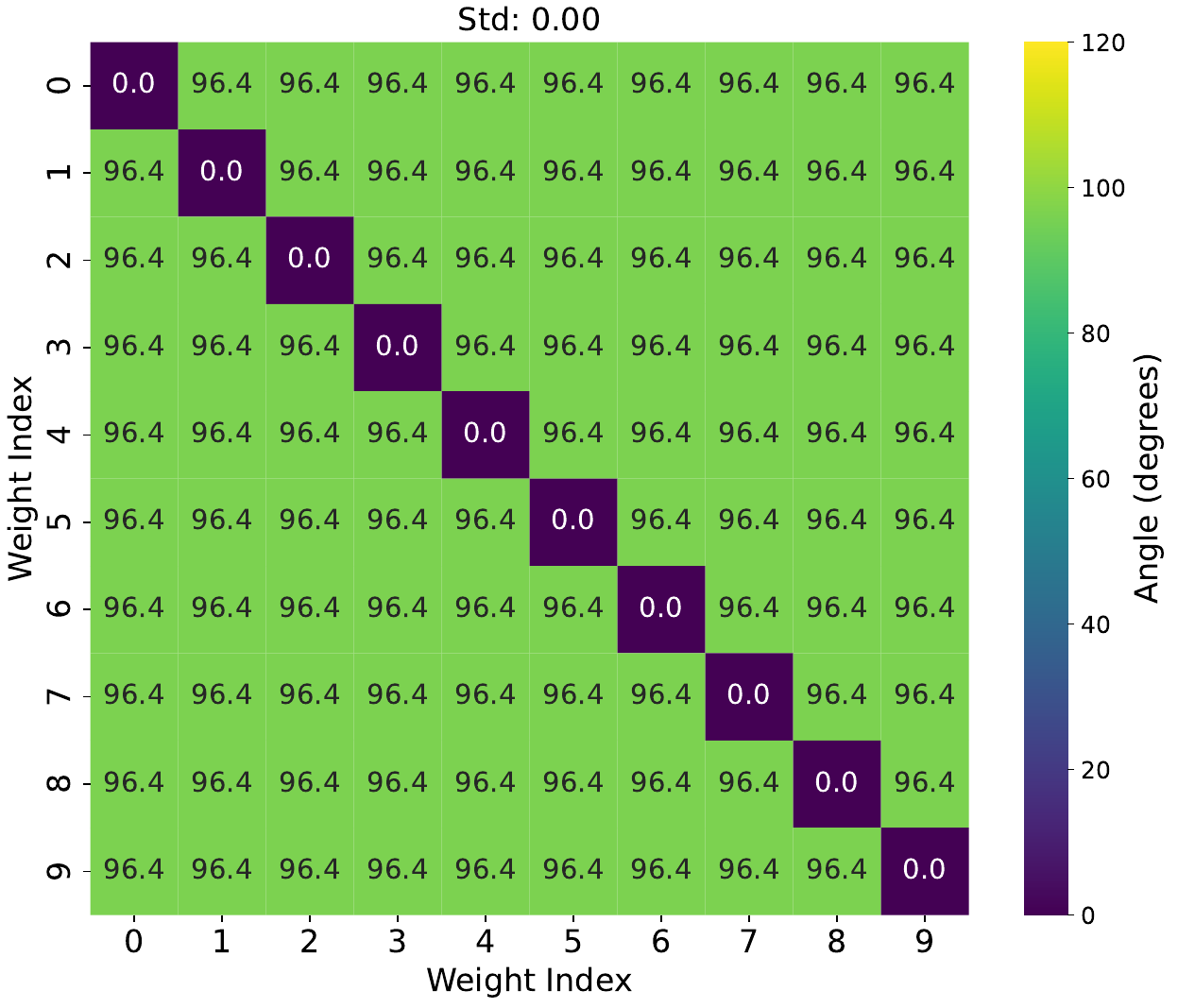}
        \caption{weight angles}
        \label{fig:etf-weight}
    \end{subfigure}
    \caption{The weights of LT-DARTS. (a) is the weight norm during the search process and (b) represents the angle between pairwise weight vectors after the search is completed.}
    \label{fig:etf-eff}
\end{figure}

We evaluate the role of the BFC in our method. It comes as no surprise that BFC ensures fairness throughout the search process. Fig.~\ref{fig:etf-eff} illustrates changes in the classifier's weights during the architecture search process. It is obvious that the weight norms remain constant and the angles are equal, as this aligns with the derivation in Eq.~(\ref{eq:etf}).

Interestingly, we observe that the BFC appears to mitigate performance collapse~\cite{Zela2019UnderstandingAR, Chen2020StabilizingDA, Ye2022betaDARTSBR, Movahedi2022DARTSMP, Ye2023DARTSBR}, which is one of the key challenges in the DARTS field. As shown in Fig~\ref{fig:performance-cls}, the vanilla DARTS often suffers from a sharp performance decline in the later stages of architecture search, with architectures identified at these stages exhibiting notably inferior performance compared to those found earlier (see the green line in the figure). However, our approach ensures continuous improvement in architecture performance throughout the search process. Previous studies indicate that a smaller $\lambda_{max}^{\alpha}$, which denotes dominant eigenvalue of  architecture parameter $\alpha$, can effectively mitigate performance collapse~\cite{Zela2019UnderstandingAR, Chen2020StabilizingDA}. Hence, we track the $\lambda_{max}^{\alpha}$ during the search process. As shown in Fig.~\ref{fig:ev}, the BFC achieves a smaller $\lambda_{max}^{\alpha}$ than the original method, providing compelling evidence that our approach successfully mitigates performance degradation and leading to improved overall search performance.

\subsection{Ablation Study}
We conduct ablation experiments to validate the improvements in both the search space and search strategy. We take the vanilla DARTS as the baseline and first improve its search space by adding two novel convolution operations, namely LT-AggConv and LT-HierConv (LT-Conv). This modification enhances the architecture's adaptability to tailed classes. Subsequently, we further improve the method by incorporating the BFC, aiming to reduce the negative impact of biased gradient updates during architecture search. The experimental results in Table~\ref{abla} show that each component contributes positively to the final performance. 
\begin{table}[ht]
\centering
\caption{Ablation studies on CIFAR-10-LT.}
\resizebox{0.65\linewidth}{!}{\begin{tabular}{c|c|c|c}
\hline
Vanilla DARTS & LT-Conv.        &    BFC   & Acc. \\ \hline
$ \surd $   &                &           & 65.3 \\
$ \surd $   & $ \surd $      &           & 74.3 \\
$ \surd $   & $ \surd $      &  $\surd$  & 77.8 \\ \hline
\end{tabular}}
\label{abla}
\end{table}



\section{Conclusion}
Long-tailed (LT) data in real-world scenarios presents formidable challenges for recognition tasks. In this paper, we approach the issue from the overlooked perspective of architecture and bridge the gap between network architecture design and LT. We conduct comprehensive experiments to investigate which architectural components perform better under LT data. Based on our observations, we identify and design two high-performing candidate operations. To enable a more thorough and efficient exploration of architectural design, we propose LT-DARTS, a NAS method specifically tailored for LT scenarios. In addition to updating the search space, we introduce the balanced fixed classifier to address the biased search process under LT conditions. Extensive experiments demonstrate that our approach effectively designs LT-friendly architectures. Moreover, our method can integrate seamlessly with existing techniques and easily surpass state-of-the-art results, offering an orthogonal solution to LT challenges.

\begin{acks}
This work was supported by the National Natural Science Foundation of China under Grant 62471451. The authors would like to thank the Information Science Laboratory Center of USTC for the hardware and software services.
\end{acks}

\section*{GenAI Usage Disclosure}
The authors used a LLM to assist with grammar checking and language polishing in the manuscript. The model was not used for generating technical content, code, data, or figures. All text was authored and fully reviewed by the human authors. The LLM served only as a writing aid, and no part of the final content is directly generated by AI.

\bibliographystyle{ACM-Reference-Format}
\bibliography{sample-base}


\end{document}